\documentclass[sigconf]{acmart}
\usepackage{algorithm}
\usepackage{algorithmic}
\usepackage{graphicx}
\usepackage{epstopdf}
\usepackage{subfigure}
\usepackage{soul}
\usepackage{multicol}
\usepackage{multirow}

\AtBeginDocument{%
  \providecommand\BibTeX{{%
    \normalfont B\kern-0.5em{\scshape i\kern-0.25em b}\kern-0.8em\TeX}}}

\copyrightyear{2021}
\acmYear{2021} 
\setcopyright{acmcopyright}
\acmConference[CIKM '21]{Proceedings of the 30th ACM International Conference on Information and Knowledge Management}{November 1--5, 2021}{Virtual Event, QLD, Australia} 
\acmBooktitle{Proceedings of the 30th ACM International Conference on Information and Knowledge Management (CIKM '21), November 1--5, 2021, Virtual Event, QLD, Australia} 
\acmPrice{15.00} 
\acmDOI{10.1145/3459637.3482393} 
\acmISBN{978-1-4503-8446-9/21/11}

\begin{document}

\title{Single Node Injection Attack against Graph Neural Networks }


\author{Shuchang Tao$^{1,3}$, Qi Cao$^{1}$, Huawei Shen$^{1,3,*}$, Junjie Huang$^{1,3}$, Yunfan Wu$^{1,3}$, Xueqi Cheng$^{2,3}$}
\affiliation{
  \institution{$^{1}$Data Intelligence System Research Center,  \\  Institute of Computing Technology, Chinese Academy of Sciences, Beijing, China\\
  $^{2}$CAS Key Laboratory of Network Data Science and Technology, }
  \country{Institute of Computing Technology, Chinese Academy of Sciences, Beijing, China\\
  $^{3}$University of Chinese Academy of Sciences, Beijing, China
}
}
\email{{taoshuchang18z,caoqi,shenhuawei,huangjunjie17s,wuyunfan19b,cxq}@ict.ac.cn}


\begin{abstract}
Node injection attack on Graph Neural Networks (GNNs) is an emerging and practical attack scenario that the attacker injects malicious nodes rather than modifying original nodes or edges to affect the performance of GNNs.  However, existing node injection attacks ignore extremely limited scenarios, namely the injected nodes might be excessive such that they may be perceptible to the target GNN. In this paper, we focus on an extremely limited scenario of \emph{single node injection evasion attack}, i.e., the attacker is only allowed to inject one single node during the test phase to hurt GNN's performance. The discreteness of network structure and the coupling effect between network structure and node features bring great challenges to this extremely limited scenario. We first propose an optimization-based method to explore the performance upper bound of single node injection evasion attack. Experimental results show that 100\%, 98.60\%, and 94.98\% nodes on three public datasets are successfully attacked even when only injecting one node with one edge, confirming the feasibility of single node injection evasion attack. However, such an optimization-based method needs to be re-optimized for each attack, which is computationally unbearable. To solve the dilemma, we further propose a \emph{\underline{G}eneralizable \underline{N}ode \underline{I}njection \underline{A}ttack model}, namely G-NIA, to improve the attack efficiency while ensuring the attack performance. Experiments are conducted across three well-known GNNs. Our proposed G-NIA significantly outperforms state-of-the-art baselines and is 500 times faster than the optimization-based method when inferring.
\let\thefootnote\relax\footnotetext{*Corresponding Author}
\end{abstract}

\ccsdesc[300]{Information systems~Data mining}

\keywords{Node injection attack; 
Adversarial attack;
Graph neural networks}

\fancyhead{}
\maketitle
{\fontsize{8pt}{8pt} \selectfont
\textbf{ACM Reference Format:}\\
Shuchang Tao, Qi Cao, Huawei Shen, Junjie Huang, Yunfan Wu, Xueqi Cheng. 2021. Single Node Injection Attack against Graph Neural Networks.  In {\it Proceedings of the 30th ACM International Conference on Information and Knowledge Management (CIKM '21), November 1–5, 2021, Virtual Event, Australia.} ACM, New York, NY, USA, 10 pages. https://doi.org/10.1145/3459637.3\\
482393}

\section{Introduction}

Graph Neural Networks (GNNs) have achieved remarkable performance in many graph mining tasks such as node classification~\cite{kipf2017semi, velickovic2018graph, hamilton2017inductive, xu2018gwnn, xu2019graphheat}, graph classification~\cite{rieck2019persistent}, cascade prediction~\cite{Cao2020PopularityPO,Qiu2018DeepInfSI} and  recommendation systems~\cite{fan2019graph}. 
Most GNNs follow a message passing scheme~\cite{wang2020amgcn}, 
learning the representation of a node by aggregating representations from its neighbors iteratively.
Despite their success, GNNs are proved to be vulnerable to adversarial attacks~\cite{Dai2018AdversarialAO,zugner2018adversarial,Bojchevski2018AdversarialAO}, which are attracting increasing research interests. Imperceptible perturbations on graph structure or node attributes can easily fool GNNs both at training time (poisoning attacks) as well as inference time (evasion attacks)~\cite{zugner_adversarial_2019}.

\begin{figure}
	\includegraphics[width=7.2cm]{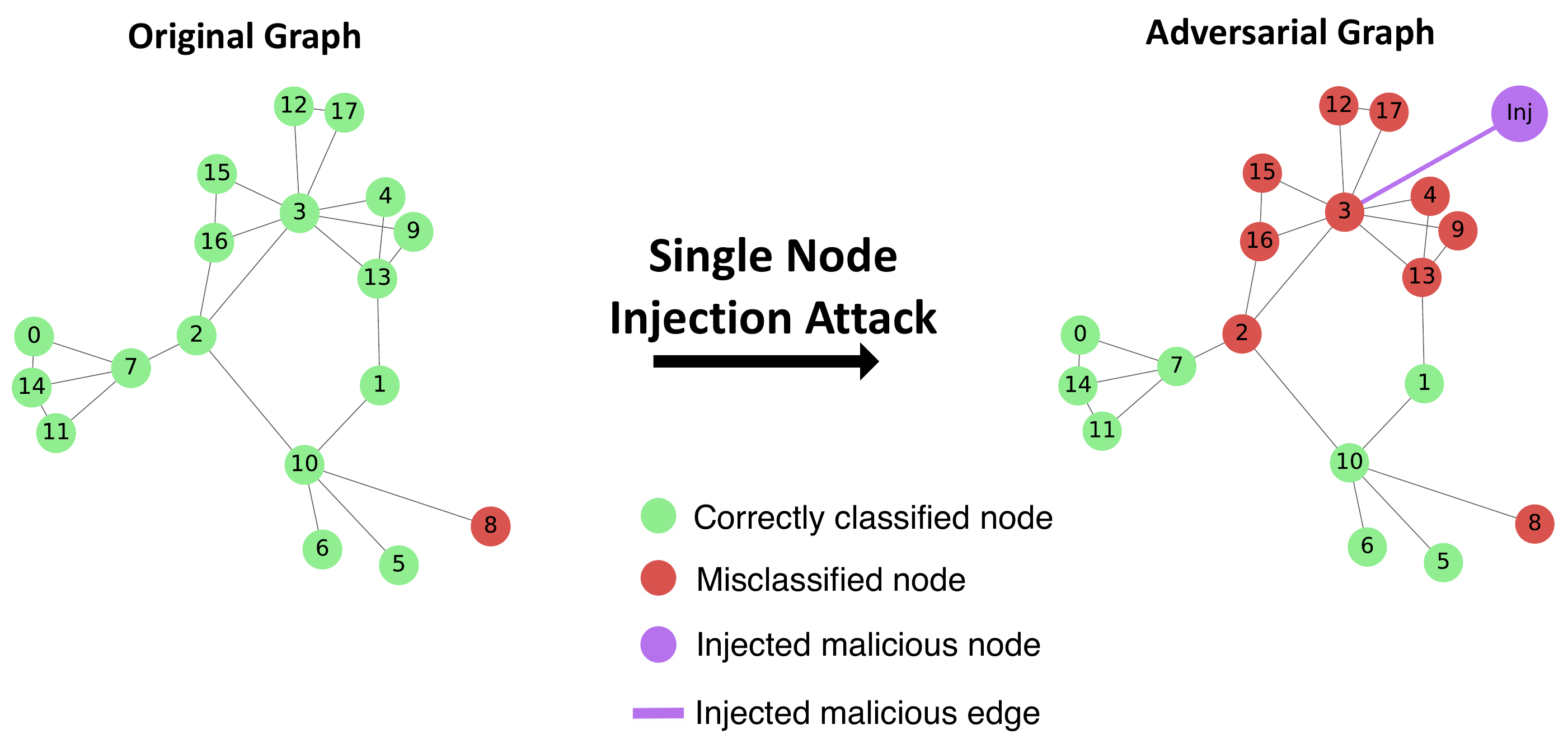}
	\caption{Comparison before and after single node injection evasion attack on Reddit. 
}
	\label{fig:example}
	\vspace{-10pt}
\end{figure}
 
Pioneer attack methods for GNNs generally modify the structure~\cite{zugner_adversarial_2019,zugner2018adversarial, Lin2020ExploratoryAA} or attributes~\cite{Finkelshtein2020SingleNodeAF, zugner2018adversarial} of existing nodes in the original graph, which requires high authority and makes the attack difficult to apply. 
The emerging node injection attack focuses on a more practical scenario. 
However, existing node injection attacks ignore extremely limited scenarios for adversarial attacks, namely the injected nodes might be excessive such that they may be perceptible or detected by both the target victim system or the target nodes. 
In addition, existing node injection attacks generally focus on poisoning scenarios, i.e., assuming that the target GNN is trained/retrained on the injected malicious nodes with labels. 
Such assumption is hard to satisfy in real scenarios since we may have no ideas about the label standard of target victim systems.

In this paper, we focus on a practical and extremely limited scenario of \emph{single node injection evasion attack}, i.e., the attacker is only allowed to inject one single node during the test phase to hurt the performance of GNNs.
Single node injection evasion attack has the advantages of being difficult to detect and low in cost.
Taking social networks as an example, it is very difficult for both the detection system and the user to detect the addition of one single malicious follower, especially compared with existing node injection methods~\cite{Sun2020AdversarialAO} which lead to a sudden emergence of multiple followers.
In addition, the focused evasion attack scenario does
 not require the vicious nodes to be present in the training set, nor does it force GNN to train on the unknown label, making it more realistic than the poisoning attack.

The key of single node injection evasion attack is to spread
malicious attributes to the proper nodes along with the network structure. 
However, the discreteness of network structure and the coupling effect between network structures and node features bring great challenges to this extremely limited scenario. 
We first propose an optimization-based method, namely OPTI, to explore the performance upper bound of single node injection evasion attack. 
To directly optimize the continuous/discrete attributes and the discrete edges of the malicious node, we extend the Gumbel-Top-$k$ technique to solve this problem. Experimental results show that OPTI can successfully attack 100\%, 98.60\%, 94.98\% nodes on three public datasets, confirming the feasibility of single node evasion injection attack.
Nevertheless, OPTI needs to be re-optimized for each attack, which is computationally unbearable.
In addition, the information learned in each optimization is discarded, which hinders the generalization of the injection attack.

To solve the dilemma, we further propose a \emph{\underline{G}eneralizable \underline{N}ode \underline{I}njection \underline{A}ttack model}, namely G-NIA, to improve the attack efficiency while ensuring the attack performance.
G-NIA generates the discrete edges also by Gumbel-Top-$k$ following OPTI and captures the coupling effect between network structure and node features by a sophisticated designed model. Comparing with OPTI, G-NIA shows advantages in the following two aspects.
One is the representation ability. G-NIA explicitly models the most critical feature propagation via jointly modeling. Specifically, the malicious attributes are adopted to guide the generation of edges, modeling the influence of attributes and edges better.
The other is the high efficiency. G-NIA adopts a model-based framework, utilizing useful information of attacking during model training, as well as saving a lot of computational cost during inference without re-optimization.

To provide an intuitive understanding of single node evasion injection attack, 
we give a real example about the attack performance of our proposed generalizable node injection attack model in Figure~\ref{fig:example}. 
Specifically, we construct a subgraph of the social network Reddit. 
The attacker makes a total of 9 nodes misclassified even in an extremely limited scenario, i.e., injecting one node and one edge.
Indeed, we observe remarkable performances when applying our proposed G-NIA model on real large-scale networks. 
Experimental results on Reddit and ogbn-products show that our G-NIA achieves the misclassification rate of 99.9\%, 98.8\%, when only injecting one node and one edge. The results reveal that G-NIA significantly outperforms state-of-the-art methods  on all the GNN models and network datasets, and is much more efficient than OPTI.
The results demonstrate the representation ability and high efficiency of G-NIA.

To sum up, our proposal owns the following main advantages:
\vspace{-4pt}
\begin{enumerate}
	\item \emph{Effectiveness:} When only injecting one single node and one single edge,
	the proposed OPTI can successfully attack 100$\%$, 98.60$\%$, 94.98$\%$ nodes on a social network, a co-purchasing network, and a citation network. 
	Our G-NIA can achieve comparable results during inferring without re-optimization like OPTI.
	Both of them significantly outperform state-of-the-art methods on all three datasets.
	\item \emph{Efficiency:}
	The proposed model G-NIA is approximately 500 times faster than OPTI in the inference phase.
	\item \emph{Flexibility:} Extensive experiments are conducted on three kinds of network datasets across three well-known GNN models in two scenarios, to demonstrate the effectiveness of our proposed G-NIA  model. In addition, G-NIA is also capable in the black-box scenario.

\end{enumerate}

\vspace{-6pt}
\section{Related Work}

\subsection{Adversarial Attacks on Graphs}
GNNs have shown exciting results on many graph mining tasks, e.g., node classification~\cite{Klicpera2018PredictTP,hamilton2017inductive}, network representation learning~\cite{10.1145/3097983.3098061}, graph classification~\cite{rieck2019persistent}. 
However, they are proved to be sensitive to adversarial attacks~\cite{Sun2018AdversarialAA,Jin2020AdversarialAA,Jia2020CertifiedRO,zhang2019comparing,Bojchevski2018AdversarialAO,zugner_adversarial_2019,Ma2021GraphRewir}.
Attack methods can perturb both the graph structure, node attributes, and labels~\cite{Chen2020ASO,tao2021advimmune,XuAADIG}.
Specifically, Nettack~\cite{zugner2018adversarial} modifies node attributes and the graph structure guided by the gradient.
RL-S2V~\cite{Dai2018AdversarialAO} uses reinforcement learning to flip edges.
Others make perturbations via approximation techniques~\cite{Wang2019AttackingGC}, label flipping~\cite{ZhangHSW20}, and exploratory strategy~\cite{Lin2020ExploratoryAA}. 

In the real world, modifying existing edges or attributes is intractable, due to no access to the database storing the graph data~\cite{Sun2020AdversarialAO}.
Node injection attack~\cite{Sun2020AdversarialAO, Wang2020ScalableAO} focuses on a more practical scenario, which only injects some malicious nodes without modifying the existing graph.
Pioneering researches focus on node injection poisoning attacks and have some problematic disadvantages which cannot be overlooked in an evasion attack scenario. 
Node Injection Poisoning Attack (NIPA) uses a hierarchical Q-learning network to sequentially generate the labels and edges of the malicious nodes~\cite{Sun2020AdversarialAO}.
Nevertheless, NIPA fails to generate the attributes of injected nodes, which are extremely important in evasion attacks, resulting in its poor performance.
Approximate Fast Gradient Sign method (AFGSM) provides an approximate closed-form attack solution to GCN with a lower time cost~\cite{Wang2020ScalableAO}, but AFGSM is difficult to handle other GNNs, which limits its application. 
In addition, AFGSM cannot deal with the continuous attributed graphs which are common in the real world.
Such approximation also leads to an underutilization of node attributes and structure.

There is still a lack of an effective method that simultaneously considers the attributes and structure of the injected node.

\vspace{-2pt}
\subsection{Attack in the Extremely Limited Scenario}

Extremely Limited Scenario is firstly proposed in one-pixel attack by Su et al.~\cite{Su2019OnePA} in the computer vision area.
The attacker can only modify one pixel to make the image misclassified.
Su et al. propose a method for generating one-pixel adversarial perturbations based on differential evolution. 
This work  starts the discussion of extremely limited scenario in adversarial learning area~\cite{Akhtar2018ThreatOA,Vargas2020UnderstandingTO,Kgler2018ExploringAE,Finkelshtein2020SingleNodeAF}.

Recently, in the graph adversarial learning area, Finkelshtein et al.~\cite{Finkelshtein2020SingleNodeAF} propose the single-node attack to show that GNNs are vulnerable to the extremely limited scenario.
The attacker aims to make the target node misclassified by modifying the attributes of a selected node.
However, it is not realistic because modifying an existing node requires high authority and makes it difficult to apply.
Up to now, there is still a lack of a practical method conducted in the extremely limited scenario in graph adversarial learning.

Our work fills the two gaps by proposing a single node injection evasion attack scenario and proposing G-NIA to ensure the effectiveness and efficiency of the model at the same time.

\vspace{-5pt}

\section{Preliminaries}
\label{sec:Pre}
This section introduces the task of node classification, together with GNNs to tackle this task.
\subsection{GNN for Node Classification}
\textbf{Node classification}
Given an attributed graph $G=(V, E, X)$, $V =\{1, 2, ..., n\}$ is the set of $n$ nodes, $E \subseteq V \times V$ is the edge set, and 
$X \in \mathbb{R}^{n \times d}$ is the attribute matrix with $d$-dimensional features. 
We denote network structure as adjacency matrix $A$.
Given $V_l \subseteq  V$ labelled by a class set $\mathcal{K}$,
the goal is to assign a label for each unlabelled node by the classifier $\boldsymbol{Z} = f_{\theta}(G) $, where $\boldsymbol{Z} \in \mathbb{R}^{n \times K}$ is the probability distribution, and $K=|\mathcal{K}|$ is the number of classes.
\textbf{Graph neural networks.}
A typical GNN layer contains a feature transformation and an aggregation operation~\cite{Ma2020AUV,You2021IDGNN,Xu2019HowPA} as:
\begin{equation}
X_{i n}^{\prime}=f_{trans}\left(X_{i n}\right); \ \
X_{out}=f_{agg}\left(X_{i n}^{\prime} ; G\right)
\end{equation}
where $X_{in} \in \mathbb{R}^{n \times d_{in}}$ and $X_{out} \in \mathbb{R}^{n \times d_{out}}$ are the input and output features of a GNN layer with $d_{in}$ and $d_{out}$ dimensions. The feature transformation $f_{trans}( \cdot )$ transforms $X_{in}$ to $X_{in}^{\prime} \in \mathbb{R}^{n \times d_{out}}$; and the feature aggregation $f_{agg}\left(\cdot\ ; G\right)$ updates features by aggregating transformed features via $G$.
Different GNNs have varied definitions of $f_{trans}( \cdot )$ and $f_{agg}\left(\cdot\ ; G\right)$.
Taking the GCN as an example: 
$
X_{in}^{\prime}=X_{in} \boldsymbol{W} ; \
X_{out}=\tilde{D}^{-\frac{1}{2}} \tilde{A} \tilde{D}^{-\frac{1}{2}} X_{in}^{\prime} ,
$
where $\tilde{A}=A +I $, $\tilde{D}$ is the degree matrix, $\boldsymbol{W} \in \mathbb{R}^{d_{in} \times d_{o u t}}$ is a weight matrix for feature transformation.

\subsection{Problem Definition}
Single node injection evasion attack aims to make the original nodes misclassified in an extremely limited scenario, i.e., the attacker is only allowed to inject one single node in the test phase without modifying the existing nodes and edges in the original graph.  
We formalize the objective function as: 
\begin{equation}
\begin{aligned}
\label{eq:loss}
\max_{G^{\prime}} \quad & \sum_{t\in V_{tar}} \mathbb{I}\left( f_{\theta^{*}} \left(G^{\prime},t \right) \neq y_t\right)  \\
G^{\prime} =& \left(V \cup \boldsymbol{v}_{inj}, E \cup \boldsymbol{e}_{inj}, X \oplus \boldsymbol{a}_{inj}\right) \\
s.t. \quad 
&\theta^{*}=\arg \min _{\theta} \mathcal{L}_{\operatorname{train}}\left(f_{\theta}(G),y\right),
\end{aligned}
\end{equation}
where $\mathbb{I}(\cdot)$ is an indicator function, $V_{tar}$ is the target nodes which can be a single node as well as multiple nodes, $y_t$ is the ground truth label of node $t$, and $f_{\theta^{*}}$ is the attacked GNN model which has been trained on the original graph $G$. The perturbed graph $G^{\prime}$  includes the malicious injected node $\boldsymbol{v}_{inj}$, together with its edges $\boldsymbol{e}_{inj}$ and attributes $\boldsymbol{a}_{inj}$. $\oplus$ refers to the concatenation operation.
To ensure the imperceptibility, we define an \textit{injected edge budget} i.e., $\|\boldsymbol{e}_{inj}\|<\Delta$.
And the malicious attributes are also limited by the maximum and minimum values of the existing attributes for continuous attributed graphs, and the maximum $L^0$ norm for discrete ones.

Node injection evasion attack takes the advantage of the feature propagation to spread the malicious attributes to the proper nodes without retraining. 
Feature propagation depends on the coupling effect of attributes and edges. Therefore, the \textbf{core} is to generate proper attributes and edges of the injected nodes, while considering their coupling effect between each other. 
Nevertheless, generating attributes and edges faces their respective challenges.

\begin{figure}[t]
	\includegraphics[width=5.3cm]{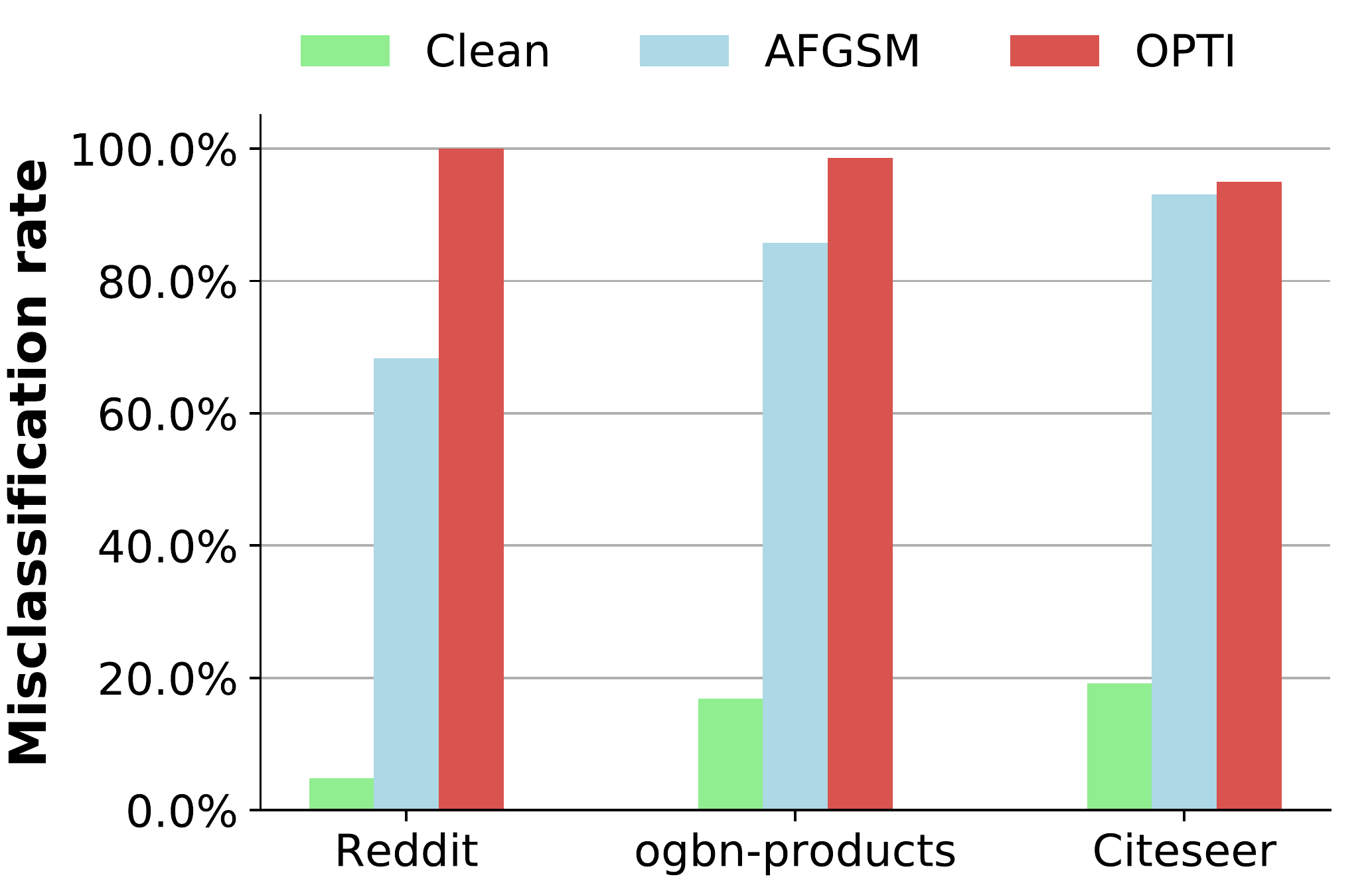}
	\caption{Misclassification rate of single node injection attack on GCN on three datasets. Clean refers to the clean graph. AFGSM is the state-of-the-art node injection attack method. OPTI is our proposed jointly optimization method. }
	\label{fig:opti}
	\vspace{-15pt}
\end{figure}
\textbf{Challenges:}
For edges, the challenge comes from the discrete nature of edges. The resulting non-differentiability hinders back-propagating gradients to guide the optimization of injected edges. 
For attributes, different network datasets vary greatly. 
Most networks, such as social networks and co-purchasing networks, have continuous attributes to denote more meanings,
while commonly used citation networks in researches have discrete node attributes to represent the appearance of keywords.
But so far, few studies have focused on continuous attribute graphs
Also, discrete attribute generation faces the same challenge as edge generation. 
It is worth noting that no node injection attack method can handle both two kinds of attributed graphs.

\begin{figure*}[t]
	\includegraphics[width=13cm]{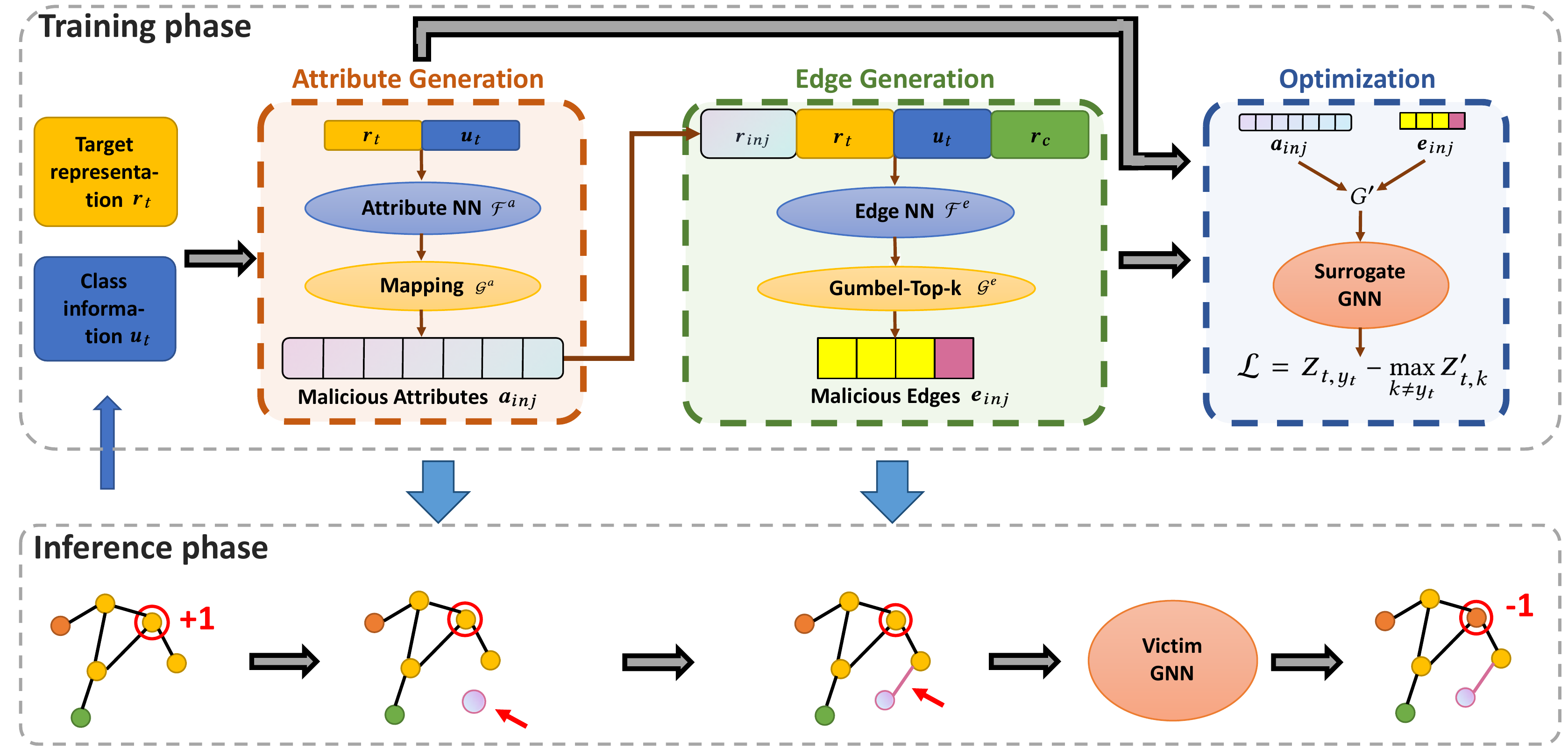}
	\caption{The overall architecture of our G-NIA model}
	\label{fig:model}
	\vspace{-6pt}
\end{figure*}

\section{Optimization-based Node Injection Attack}
\label{sec:opti}
In this section, we first propose an optimization-based approach, which is simple but effective. 
We evaluate it on three well-known datasets.
Experimental results indicate that this method explores the performance upper bound.
These results prove that the single node injection attack is sufficient for the targeted evasion attack, and it is time to explore the efficiency of the attack.

\subsection{Optimization-based Approach}
To tackle the single node injection evasion attack problem, 
we propose an optimization-based approach, namely OPTI, to directly optimize the attributes and edges of the vicious node. 
Specifically, we first initialize the vicious attributes and edges as free variables and inject them into the original graph.
Then, we feed the perturbed graph into a surrogate GNN model to obtain the output.
After that, we calculate the loss by Equation~\ref{eq:loss}  and optimize the free variables until convergence. 
The joint optimization helps to depict the coupling effect between the attributes and edges of the vicious node.
To overcome the gradient challenge of discrete variables as we mentioned above, we adopt the Gumbel-Top-$k$ technique to aid in optimizing, which will be specifically introduced in Section~\ref{sec:gumbel}.

\subsection{Attack Performance}
\label{sec:opti_perform}
We execute the single node injection evasion attack on three datasets.
For each attack, the attacker only injects one node and one edge to attack one target node during the inference phase.
Every node in the whole graph is selected as a target node, separately.
We will elaborate on the settings and datasets in Section~\ref{sec:settings}. Here, we just tend to illustrate the attack performance.

As shown in Figure~\ref{fig:opti}, our proposed method OPTI significantly outperforms state-of-the-art AFGSM on all three datasets.  OPTI makes 100\%, 98.60\%, 94.98\% nodes misclassified on Reddit, ogbn-products, and Citeseer, even when injecting only one single node and one single edge.
The results show that OPTI can attack almost all target nodes successfully.
It also confirms the feasibility and effectiveness of the single node injection evasion attack, thus we do not need to inject more nodes on this attack scenario.

Even though OPTI achieves excellent results, it is difficult to be executed in reality. 
Because it is a case-by-case method, which means a distinct optimization is executed for each attack. 
The optimization is computationally expensive, and no knowledge can be saved and reused.
What's worse, these limitations exist in almost all attack methods in the graph adversarial learning area.
Therefore, it is time to explore the efficiency of single node injection attack.

\vspace{-5pt}
\section{Generalizable Node Injection Attack}
In this section, we first introduce the generalization ability in the graph adversarial attack area. 
Then, we creatively propose a generalizable attack model G-NIA to carry out the node injection attack, which generates attack plans through the knowledge learning from the original graph.

\vspace{-3pt}
\subsection{Generalization in Graph Adversarial Attack}
Generalization refers to the adaptability of machine learning algorithms to fresh (unseen) samples.
In the field of graph adversarial learning, generalization is also important, especially in targeted attacks~\cite{Baluja2018LearningTA}.
For example, the above optimization-based method is almost impossible to tackle a real-life attack, even though it achieves superb attack performance. Because the huge computation brought by the case-by-case optimization is intolerable.

To solve the dilemma, we propose to model the optimization process by a generalizable model. 
The generalizable model should have the ability of both fitting and generalization.
The fitting ability refers to having high attack performance during training, fitting OPTI as much as possible. Note that the fitting ability is the fundamental condition for generalization. 
A parametric attack model with poor attack performance does not meet the standard of generalization.
The generalization ability refers to the performance and efficiency of (fresh) test data which is unseen in the training phase. 
The efficiency is naturally guaranteed by the parametric model since the knowledge has been preserved by the parameters.
Note that, unlike those one-off optimization methods, the model can be saved and reused. A well-trained model can learn the information that is implicit in the data. Thus, the model can attack successfully on unseen fresh nodes, as long as they obey the assumption of independent and identical distribution.

\vspace{-6pt}
\subsection{Framework of Generalizable Node Injection Attack Model}

We design and propose \emph{\underline{G}eneralizable \underline{N}ode \underline{I}njection \underline{A}ttack model}, namely G-NIA, to perform the single node injection evasion attack on graph data, which brings the generalization ability while keeping the high attack performance as the above OPTI.

Figure~\ref{fig:model} illustrates the overall framework of the proposed model G-NIA. 
The upper part shows the workflow of our proposed G-NIA during training. 
Specifically, G-NIA consists of three components: attribute generation, edge generation, and optimization.
The first two components generate proper attributes and edges for the malicious injected node, 
and the last one performs an attack on the surrogate GNN and calculates the loss to optimize G-NIA.
The lower part shows how G-NIA attacks the original graph in the inference phase.
G-NIA first generates the malicious attributes, then generates edges, and uses the obtained adversarial graph $G^{\prime}$ to attack the target nodes.
Next, we will provide the fine details of each component in our proposed model G-NIA.

\subsection{Attribute Generation}
\label{sec:attr}
The key point of the node injection evasion attack is that the injected node propagates the malicious attributes to proper nodes through feature aggregation. 
This process requires close cooperation between the attributes and edges of the malicious node. 
Among them, attributes are essential for attack performance, since attributes bring additional information to a successful attack. 

The goal of the attack is to make target nodes misclassified, which is inherently related to target nodes as well as  class information,
including the label class, i.e., $y_t$, and the class that is most likely to be misclassified by a surrogate GNN, i.e., $k_t$.
Specifically, target nodes provide necessary information about who we are going to attack, and the class information tells
a possible way to conduct the attack, i.e., changing their predicted labels from $y_t$ to $k_t$.
Based on these, we adopt the representation of target nodes $\boldsymbol{r}_{t}$ as well as the  representation of label class and misclassified class, i.e., $\boldsymbol{u}_t = \{\boldsymbol{u}_{y_t}, \boldsymbol{u}_{k_t}\}$, to generate attributes.
For the representation of target nodes, $\boldsymbol{r}_{t}$   is calculated by the surrogate GNN $f$.
For the representation of label class $\boldsymbol{u}_{t}$, we adopt the $y_t$-th column of the feature transformation weights, i.e., $\boldsymbol{u}_{y_t} = \boldsymbol{W}_{[:,y_t]}$ and $\boldsymbol{u}_{k_t} = \boldsymbol{W}_{[:,k_t]}$, where  $\boldsymbol{W} = \boldsymbol{W}_0 \boldsymbol{W}_1$ in a GNN with two layers.
Because feature transformation is a mapping between the attribute space and label space.
 
Feeding the above representations, we utilize two layer neural networks $\mathcal{F}^{a}$ and a mapping function $\mathcal{G}^{a}$ to generate attribute of the malicious injected node $\boldsymbol{a}_{inj}$:
\begin{equation}
\begin{aligned}
\label{eq:attr}
\boldsymbol{a}_{inj} &= \mathcal{G}^{a}\left(\mathcal{F}^{a}\left(V_{tar}, f, G; \theta^{*} \right)\right) \\
\mathcal{F}^{a} \left(V_{tar}, f, G; \theta^{*} \right) &=   \sigma \left( \left[ \boldsymbol{r}_{t} \| \boldsymbol{u}_t \right]\boldsymbol{W}_{0}^{a}   + \boldsymbol{b}_{0}^{a} \right)\boldsymbol{W}_{1}^{a} + \boldsymbol{b}_{1}^{a},
\end{aligned}
\end{equation}
where  $\theta=\left\{\boldsymbol{W}_{0}^{a},\boldsymbol{W}_{1}^{a},\boldsymbol{b}_{0}^{a},\boldsymbol{b}_{1}^{a}\right\}$ are the trainable weights. 
$\mathcal{G}^{a}$ maps the output of $\mathcal{F}^{a}$ to the designated attributes space of the original graph, making the attributes similar to existing nodes.
For continuous attributes, $\mathcal{G}^{a}$ contains a sigmoid function and a scaler to stretch to the designated attributes space. 
For discrete attributes, $\mathcal{G}^{a}$ is the Gumbel-Top-$k$ technique, which will be introduced in Section~\ref{sec:gumbel}.

Comparing with other approaches, G-NIA is more flexible.
Our model can handle high-dimensional attributes generation for both discrete and continuous attributed graphs,
but AFGSM can only deal with discrete attributes and NIPA even cannot generate attributes.

\subsection{Edge Generation}
\label{sec:edge}

The malicious edges help the malicious node to spread its attributes to the required candidate nodes.
Here, the candidate nodes are limited within the target nodes and their first-order neighbors, making the injected node to be at least the second-order neighbors of the target nodes. 
Because most GNNs with good performance contain two layers of feature propagation. 
Owing to the limited injected edge budget $\Delta$, we score the candidate nodes and select the best $\Delta$ ones to connect to the injected node.
The scores of candidate nodes measures the impact of connecting certain candidate nodes to the malicious node on the attack performance.

\subsubsection{Jointly modeling.}
To capture the coupling effect between network structure and node features, we jointly model the malicious attributes and edges.
Specifically,
we use malicious attributes to guide the generation of malicious edges, \textbf{jointly modeling} the attributes and edges of the injected node. 
According to this, we use the representation of candidate nodes $\boldsymbol{r}_c$ and the representation of the malicious node $\boldsymbol{r}_{inj}$  to obtain scores.
Specifically, $\boldsymbol{r}_{inj}$ is computed by feature transformation of the surrogate GNN, i.e., $\boldsymbol{r}_{inj} = f_{trans}(\boldsymbol{a}_{inj})$, since the malicious node does not have structure yet. 
(The brown arrow from malicious attributes to the input of edge generation in Figure~\ref{fig:model}.)

As we mentioned before, the goal of the attack is closely tied to two elements, target nodes, and class information.
The representations of the above two elements, i.e. $\boldsymbol{r}_{t}$ and $\boldsymbol{u}_{t}$, are also included in the input of neural networks, and they are both repeated $m$ times, where $m$ is the number of candidate nodes.

With all the mentioned inputs, we adopt two layer neural networks $\mathcal{F}^{e}$ to obtain the score of candidate nodes, and use Gumbel-Top-$k$ technique $\mathcal{G}^{e}$ to discretize the scores as follows:
\begin{equation}
\begin{aligned}
\label{eq:edge}
& \boldsymbol{e}_{inj} = \mathcal{G}^{e}\left(\mathcal{F}^{e}\left(\boldsymbol{a}_{inj}, V_{tar}, f, G; \theta^{*} \right)\right)
\end{aligned}
\vspace{-3pt}
\end{equation}
\begin{equation*}
\begin{aligned}
\label{eq:edge}
& \mathcal{F}^{e} \left(\boldsymbol{a}_{inj}, V_{tar}, f, G; \theta^{*} \right) = \sigma \left(\left[ \boldsymbol{r}_{inj} \| \boldsymbol{r}_{t} \| \boldsymbol{u}_{t} \| \boldsymbol{r}_{c} \right] \boldsymbol{W}_{0}^{e}  + \boldsymbol{b}_{0}^{e} \right)\boldsymbol{W}_{1}^{e}  + \boldsymbol{b}_{1}^{e},
\end{aligned}
\end{equation*}
where $\theta=\left\{\boldsymbol{W}_{0}^{e},\boldsymbol{W}_{1}^{e},\boldsymbol{b}_{0}^{e},\boldsymbol{b}_{1}^{e}\right\}$ are the trainable weights.

\subsubsection{Gumbel-Top-$k$ technique}
\label{sec:gumbel}
To tackle the discreteness of network structure, we adopt the Gumbel-Top-$k$ technique to solve the optimization of edges.  
Gumbel distribution $\mathbf{G}_{i}$ is used to model the distribution of the extremum of a number of samples of various distributions~\cite{gumbel1954statistical,Efraimidis2006WeightedRS}. Formally, $\mathbf{G}_{i}=-\log \left(-\log \left(U_{i}\right)\right)$ where $U_{i} \sim \text{Uniform}(0,1)$ is Uniform distribution. 
Gumbel-Softmax is:
\begin{equation}
\operatorname{Gumbel-Softmax}(\boldsymbol{z})_i= \frac{\exp\left(\frac{\boldsymbol{z}_{i} +\mathbf{G}_{i}}{\tau}\right)}{\sum_{j=1}^{n} \exp\left(\frac{\boldsymbol{z}_{j} +\mathbf{G}_{j}}{\tau}\right)},
\end{equation}
where $\tau$ is the temperature controlling the smoothness, $\boldsymbol{z}$ is the output of the neural network $\mathcal{F}^{e}$.
Gumbel-Softmax brings exploration to the edge selection through Gumbel distribution $\mathbf{G}_{i}$.
To encourage exploring, we further use $\epsilon$ to control the strength of exploration:
\begin{equation}
\operatorname{Gumbel-Softmax}(\boldsymbol{z}, \epsilon)_i = \frac{\exp\left(\frac{\boldsymbol{z}_{i} +\epsilon \mathbf{G}_{i}}{\tau}\right)}{\sum_{j=1}^{n} \exp\left(\frac{\boldsymbol{z}_{j} +\epsilon \mathbf{G}_{j}}{\tau}\right)}.
\end{equation}
Gumbel-Top-$k$ function $\mathcal{G}^{e}$ is:
\begin{equation}
\mathcal{G}^{e}(\boldsymbol{z}) =\sum_{j=1}^{k} \operatorname{Gumbel-Softmax}(\boldsymbol{z} \odot {mask}_j, \epsilon),
\vspace{-2pt}
\end{equation}
where $k$ is the budget of edges (discrete attributes), ${mask}_j$ filters the selected edges/attributes to ensure they will not be selected again. 
Note that the obtained vector is sharp but not completely discrete, which benefits training.
In the test phase, we discretize the vector compulsorily to obtain discrete edges.
And we do the same for discrete attributes.

Gumbel-Top-$k$ technique can solve the optimization problem of high-dimensional discrete attributes. But it is difficult for reinforcement learning methods to deal with huge discrete action spaces.   
It may be the reason why NIPA~\cite{Sun2020AdversarialAO} cannot generate attributes.
In addition, Gumbel-Top-$k$ does not affect training stability.

\subsection{Optimization}
\label{sec:loss}

After generating the malicious attributes and edges, we inject the malicious node into the original graph $G$ to obtain the perturbed graph $G^\prime$.
We feed $G^\prime$ into the surrogate GNN model and compute the attack loss $\mathcal{L}_{atk}$:
\begin{equation}
\begin{aligned}
\label{eq:loss}
\min_{G^{\prime}} \quad 
&\mathcal{L}_{a t k}\left(f_{\theta^{*}}\left(G^{\prime}\right),V_{tar}\right)= \sum_{t\in V_{tar}} \left( \boldsymbol{Z}_{t, y_t}^{\prime}-\max _{k \neq y_t} \boldsymbol{Z}_{t, k}^{\prime} \right). \\
\end{aligned}
\end{equation}
Attack loss $\mathcal{L}_{a t k}$ is used to guide the training process.
We iteratively optimize the attack loss by gradient descent until convergence.

\section{Experiments}

\subsection{Experimental Settings}
\label{sec:settings}
\subsubsection{Datasets}
To illustrate the wide adaptability of the proposed G-NIA to different application scenarios,
we conduct experiments on three different types of network datasets: a social network Reddit~\cite{Zeng2019GraphSAINTGS, hamilton2017inductive}, a products co-purchasing network ogbn-products~\cite{Hu2020OGB} and a commonly used citation network Citeseer ~\cite{Bojchevski2019CertifiableRT}.
Specifically,
in Reddit, each node represents a post, with word vectors as attributes and community as the label, while each edge represents the post-to-post relationship.
In ogbn-products, a node represents a product sold in Amazon with the word vectors of product descriptions as attributes and the product category as the label, and edges between two products indicate that the products are purchased together. 
In Citeseer, a node represents a paper with sparse bag-of-words as attributes and paper class as the label, and the edge represents the citation relationship.  
Note that the attributes of the first two datasets are continuous and Citeseer has discrete attributes.
G-NIA can effectively attack both types of attributed graphs.

Due to the high complexity of some GNN models, it is difficult to apply them to very large graphs with more than 200k nodes.
Thus, we keep the subgraphs of Reddit and ogbn-products for experiments. 
For Reddit, we randomly sample 12k nodes as a subgraph and then select the largest connected components (LCC) of the subgraph. 
For ogbn-products, we randomly sample 55k nodes as a subgraph and then also keep the LCC.
Following the settings of most previous attack methods~\cite{Wang2020ScalableAO,Sun2020AdversarialAO,zugner2018adversarial,zugner_adversarial_2019}, experiments are conducted on the LCC for all the three network datasets. The statistics of each dataset are summarized in Table~\ref{tab:dataset}.

 \begin{table}[t]
  \caption{Statistics of the evaluation datasets}
  \label{tab:dataset}
  \resizebox{8.5cm}{9.5mm}{
  \begin{tabular}{l|c|ccccc}
    \toprule
    Dataset & Type & $N_{LCC}$  & $d$ & $K$ & $D_{avg}$ \\
    \midrule
    Reddit & Social network & 10,004& 602 & 41 &  3.7\\
    ogbn-products & Co-purchasing network & 10,494 & 100 & 35 & 7.4 \\
    Citeseer & Citation network & 2,110  & 3,703 & 6 & 4.5\\
  \bottomrule
\end{tabular}}
  \vspace{-5pt}
\end{table}

\begin{table*}[]
\caption{Misclassification rate of Single Target Attack across various GNNs on Reddit, ogbn-products and Citeseer}
\label{tab:gcn_singleedge}
\begin{tabular}{l|c|c|c|c|c|c|c|c|c}
\toprule
         & \multicolumn{3}{c}{GCN}                                & \multicolumn{3}{|c|}{GAT}                                & \multicolumn{3}{c}{APPNP}                              \\
  \hline
         & Reddit           & ogbn-products    & Citeseer         & Reddit           & ogbn-products    & Citeseer         & Reddit           & ogbn-products    & Citeseer         \\
 \hline
Clean    & 8.15\%           & 21.01\%          & 22.04\%          & 8.45\%           & 24.34\%          & 21.09\%          & 6.65\%           & 18.58\%          & 19.91\%          \\
Random   & 11.33\%          & 27.65\%          & 24.05\%          & 10.45\%          & 29.60\%          & 22.23\%          & 8.05\%           & 22.44\%          & 20.83\%          \\
MostAttr  & 13.53\%          & 31.82\%          & 27.80\%          & 12.30\%          & 33.18\%          & 24.41\%          & 9.95\%           & 26.56\%          & 22.80\%          \\
PrefEdge & 8.53\%           & 23.05\%          & 22.75\%          & 8.70\%           & 26.76\%          & 21.64\%          & 6.77\%           & 20.11\%          & 20.21\%          \\
NIPA     & 10.54\%          &     22.20\%             & 22.99\%          & 10.19\%          & 25.49\%          & 21.56\%          & 9.55\%           &     20.10\%             & 19.91\%          \\
AFGSM    & 68.57\%          & 85.76\%          & 84.60\%          & 74.11\%          & 80.28\%          & 57.11\%          & 45.43\%          & 74.18\%          & \textbf{41.47\%}         \\
G-NIA    & \textbf{99.90\%} & \textbf{98.76\%} & \textbf{85.55\%} & \textbf{99.85\%} & \textbf{93.23\%} & \textbf{73.70\%} & \textbf{91.05\%} & \textbf{98.67\%} & \textbf{41.47\%} \\       
\bottomrule
\end{tabular}
\end{table*}

\subsubsection{Baseline Methods}
Since the node injection attack is an emerging type of attack, which is firstly proposed on~\cite{Sun2020AdversarialAO}, only two researches are focusing on this area. 
To demonstrate the effectiveness of our proposed model G-NIA, we design a random method and two heuristic-based methods as our baselines.
We also compare G-NIA with the two state-of-the-art node injection attack methods.

$\bullet$ \emph{Random.} 
We randomly sample a node from the whole graph and take its attributes as the attributes of the malicious node to make the attack difficult to be detected.
As for edges, we randomly select $\Delta$ nodes from the target nodes and their neighbors to ensure the injected node can influence the target node.

$\bullet$ \emph{Heuristic-based methods.} 
We design two heuristic methods from the perspective of attributes and edges.
(1) MostAttr. For the attributes of the malicious node, we use the attributes of a node randomly selected from the class that is most likely to be misclassified, i.e. $k_t$. The edge selection is the same as Random.
(2) PrefEdge. We adopt the preferential attachment mechanism~\cite{Sun2020AdversarialAO} to inject edges. Specifically, it connects the nodes with a probability proportional to the degree of the nodes, which preferentially injects edges to high-degree nodes from the targets nodes and their neighbors. The attributes are obtained by randomly sampling, the same as Random.

$\bullet$ \emph{NIPA.}
NIPA~\cite{Sun2020AdversarialAO} is a node injection poisoning attack method, which generates the label and edges of the injected node via deep reinforcement learning. Since the code is not released, we reproduce the code and adapt it to the evasion attack scenario. For attributes, it originally applies a Gaussian noise $\mathcal{N}(0, 1)$ on the average attributes of existing nodes. We concern it is too weak for the evasion attack. Hence, we compute the average of attributes on the generated label and apply a Gaussian noise $\mathcal{N}(0, 1)$ added on to them. Afterward, we discretize or rescale the output, which is the same as our method, to obtain the malicious attributes.

$\bullet$ \emph{AFGSM.}
AFGSM~\cite{Wang2020ScalableAO} is a targeted node injection attack, which calculates the approximation of the optimal closed-form solutions for attacking GCN.
It sets the edges (discrete attributes) with the largest $b$ approximate gradients to 1 where $b$ is the budget.
For continuous attributed graph, we do not limit the budget of attributes and set all the attributes with positive gradients to 1, to maximize its attack ability as much as possible.
Note that AFGSM has already outperformed existing attack methods (such as Nettack~\cite{zugner2018adversarial} and Metattack~\cite{zugner_adversarial_2019}), thus, we just compare our proposed G-NIA with the state-of-the-art AFGSM.
 
\subsubsection{Experimental Configuration}

We conduct experiments to attack across three well-known GNNs (GCN, GAT, and APPNP).
The hyper-parameters of them are the same with original works~\cite{kipf2017semi,velickovic2018graph,Klicpera2018PredictTP}.
We randomly split 80\% nodes for the training phase, and the rest 20\% for the inference phase.  
We also randomly select 20\% nodes from the training split as a validation set, which are used for hyper-parameters tuning and early stopping.
The train/validation/test split is consistent in training GNN and the proposed G-NIA, which means G-NIA first trains on the nodes of the training set as the target nodes, and then validates on the nodes of the validation set, finally evaluates the performance on the test set.

For our proposed model G-NIA, we employ RMSprop Optimizer~\cite{Graves2013GeneratingSW} to train it for at most 2000 epochs and stop training if the misclassification rate of the validation set does not increase for 100 consecutive epochs.
Learning rate is tuned over $\{1^{-5}, 1^{-4}, 1^{-3}\}$, and the temperature coefficient $\tau$ of Gumbel-Top-$k$   over $\{0.01, 0.1, ..., 100\}$.
We use exponential decay for the exploration coefficient $\epsilon$.
For the baseline methods, the settings are the same as described in the respective papers~\cite{Sun2018AdversarialAA,Wang2020ScalableAO}.

To demonstrate the capability of G-NIA, we design two scenarios for single node injection evasion attack, one is to attack a single target node, namely \textbf{Single Target Attack}, and the other is to attack multiple target nodes,  namely \textbf{Multiple Targets Attack}.
Note that existing target node injection attacks, such as AFGSM, can only attack one target node. Thus, we adjust it to adapt to the scenario of  Multiple Targets Attack.

To ensure a low attack cost, we strictly limit the injected edge budget $\Delta$ of the malicious node. 
For Single Target Attack, $\Delta=1$, which means the attacker is only allowed to inject one single edge.
For Multiple Targets Attack, the edge budget is no more than $\Delta=\min \left(n_t D_{\text{avg}}, 0.5* m \right)$, where $m$ is the number of candidate nodes, $n_t$ is the number of target nodes, $D_{\text{avg}}$ is the average degree which is shown in Table~\ref{tab:dataset}.
And the injected attributes are also forced to be consistent with the original graph. 
This change is to make the settings more realistic.
 
All experiments run on a server with Intel Xeon E5-2640 CPU, Tesla K80, and 128GB RAM, which installs Linux CentOS 7.1.

\vspace{-10pt}
\subsection{Performance of Single Target Attack}
To evaluate the effectiveness and broad applicability of our proposed G-NIA, we conduct experiments on the Single Target Attack scenario on three benchmark datasets across three representative GNNs, i.e., GCN, GAT, and APPNP.
In this scenario, the goal of the attacker is to attack one target node. 
The attacker is only allowed to inject one single node and one single edge to minimize the attack cost. 
We compare our proposed G-NIA model with the above heuristic and state-of-the-art baselines. 
The misclassification of the clean graph, i.e., Clean, is also included here as a lower bound.

Table~\ref{tab:gcn_singleedge} shows the misclassification rate in the test phase.
We can see that all the methods, including Random, increase the misclassification rate across three GNNs comparing with Clean.
The heuristic baselines perform consistently on three GNNs. MostAttr performs better than Random.
Comparing with Random, MostAttr only samples attributes from the most likely class $k_t$.
This indicates that well-designed malicious attributes benefit the attack performance. 
However, PrefEdge even leads to a decrease compared with Random and is similar to Clean. The result indicates that the preferential attachment mechanism is not suited for node injection attacks.

For the state-of-art node injection attack baselines,
NIPA performs well on the continuous attributed graphs, Reddit and ogbn-products, while on the discrete attributed graph Citeseer, it performs even similar to Random.
The result of Citeseer is different from what is shown in the original paper~\cite{Sun2020AdversarialAO} because we discretize the attributes to avoid the malicious node as outliers, which is performed for all methods.
We also evaluate our reproduced code on the original poisoning attack settings, and the results show that our code without discretizing the attributes can reproduce the attack performance in the paper. 
We believe the performance on that paper is brought by the continuous malicious attributes.
AFGSM performs best among all the baselines, especially on Citeseer dataset. This is because AFGSM can only generate discrete attributes. Its good performance indicates the effectiveness of the closed-form solution for attacking GCN proposed by AFGSM.

Our proposed model G-NIA outperforms all the baselines (or performs the same as the state-of-the-art baseline) on all datasets attacking all the three GNNs.
On continuous attributed graphs, G-NIA shows tremendous attack performance, making nearly all the nodes misclassified.
Specifically, G-NIA achieves 99.90\% and  98.76\% misclassification rate on Reddit and ogbn-products on GCN.
As for GAT and APPNP, G-NIA also achieves the misclassification rate of more than 91\%.
On the discrete attributed graph Citeseer, G-NIA still performs better than state-of-the-art AFGSM on GCN, which indicates that G-NIA learns the knowledge neglected by the approximate optimal solution of AFGSM.
Attacking GAT, G-NIA regains a significant advantage, indicating its adaptive capacity for different models.
Attacking APPNP, G-NIA performs the same as AFGSM.
This may be because that APPNP has multiple layers (more than 2 layers) of feature propagation but our malicious node selects the edge from only the one-order neighbors of the target node, making some information being neglected.
Our G-NIA shows its superiority on the continuous attributed graph, and still has room to be improved on the discrete attributed graph.

In this setting, OPTI achieves misclassification rates of 100\% on Reddit, 98.60\% on ogbn-products, and 86.97\% on Citeseer when attacking GCN.
Note that the misclassification rate in  Section~\ref{sec:opti_perform} is 94.98\% on Citeseer. The difference comes from the number of non-zero attributes of the vicious node being the maximum value of the original nodes (Section~\ref{sec:opti_perform}) or the average value (here).
Our model G-NIA performs comparably with OPTI on all three datasets without re-optimization. 
The results reveal the attack performance and generalization ability of our proposed G-NIA.

\subsection{Performance of Multiple Targets Attack}

\begin{table}[t]
\caption{Misclassification rate of Multiple Targets Attack on GCN on Reddit, ogbn-products, and Citeseer}
\label{tab:multar}
\begin{tabular}{l|c|c|c}
\toprule
         & Reddit           & ogbn-products    & Citeseer         \\
\hline
Clean    & 3.89\%           & 16.28\%          & 19.70\%          \\
Random   & 5.43\%           & 21.69\%          & 16.49\%          \\
MostAttr  & 6.40\%           & 24.08\%          & 18.19\%          \\
PrefEdge & 5.59\%           & 22.02\%          & 16.24\%          \\
NIPA     & 13.96\%          &   21.56\%               & 16.09\%          \\
AFGSM    & 51.86\%          & 87.34\%          & 63.22\%          \\
G-NIA    & \textbf{92.62\%} & \textbf{95.42\%} & \textbf{63.51\%} \\
\bottomrule
\end{tabular}
\vspace{-7pt}
\end{table}

To further demonstrate the attack ability of G-NIA, we conduct experiments on the Multiple Targets Attack scenario.
It is a difficult scenario, where the attacker is required to make all nodes in the group of multiple targets misclassified with only one injected node.
Each group contains three target nodes, and their distance is at most two hops.
It is a reasonable scenario since the attacker may inject malicious information into one community for each attack.
Note that the $\boldsymbol{r}_t$  is the average of the representation of all target nodes in each group, and so is $\boldsymbol{u}_t$
Also, we make sure there is no intersection between groups. 
We select all groups that meet the criteria from each dataset, i.e. 2959 groups in Reddit, 2620 groups in ogbn-products, 578 groups in Citeseer.
We randomly split 80\% of groups for training and the other 20\% for tests. We also randomly split 20\% from the training split as a validation set.
Owing to space limitations, we only report experimental results on GCN on three datasets.
And the OPTI is still not included due to the expensively computational cost.

The baselines show similar performance to that in the above Single Target Attack on Reddit and ogbn-products.
However, on Citeseer, the misclassification rate of heuristic baselines and NIPA decreases, comparing with Clean.
This may be because we only increase the edge budget rather than the most critical attribute budget. 
In addition, the discrete attributed graph is more difficult to attack, since the attribute budget is small.
On continuous attributed graphs, MostAttr performs better than Random, while PrefEdge shows less effectiveness.
NIPA achieves better results on the continuous attributed graph.
AFGSM performs best among baselines on all datasets.

For our model, G-NIA outperforms all the baselines on all datasets. 
G-NIA shows great attack performance, making 92.62\%, 95.42\%, and 63.51\% nodes misclassified successfully on Reddit, ogbn-products, and Citeseer.
We can see that G-NIA has a great advantage on continuous attributed graphs, and can successfully attack almost all nodes.
On the discrete attributed graph Citeseer, G-NIA still performs better than the state-of-the-art AFGSM.

\subsection{Black-box Attack}
To explore G-NIA's possibility in the real world, we execute the black-box attack. 
We adopt the most limited black-box scenario, i.e., the attacker has no access to any information of the victim GNN model (the type, parameters, and outputs of GNN).
Specifically, all the attack methods use a surrogate GCN to obtain the malicious nodes and are evaluated on completely unknown models.
Here, we use GAT and APPNP to evaluate the attack methods.
Since it is a very hard scenario, we only compare G-NIA to the state-of-the-art baselines, i.e., AFGSM and NIPA.

As shown in Table~\ref{tab:blackbox}, the performance of NIPA is still unsatisfactory.
AFGSM performs well whose misclassification rate is even 1\% higher than our model on Citeseer.
On Reddit and ogbn-products, our proposed ~G-NIA also achieves the misclassification rate of over 90\% in this black-box scenario, significantly outperforming all the baselines.
The result shows that ~G-NIA can maintain its advantages attacking unknown GNN models on continuous attributed graphs, while on discrete data, there is room for improvement.

\begin{table}[]
\caption{Black-box Attack: Misclassification rate of Single Target Attack on GAT and APPNP on Reddit, ogbn-products, and Citeseer}
\label{tab:blackbox}
\begin{tabular}{l|l|c|c|c}
\toprule
\multicolumn{1}{l}{}   &       & Reddit           & ogbn-products    & Citeseer         \\
\hline
\multirow{4}{*}{GAT}   & Clean & 8.45\%           & 24.34\%          & 21.09\%          \\
                       & NIPA  & 12.54\%          & 25.92\%          & 21.09\%          \\
                       & AFGSM & 45.83\%          & 74.85\%          & \textbf{38.63\%} \\
                       & G-NIA & \textbf{94.60\%} & \textbf{95.00\%} & 36.49\%          \\
\hline
\multirow{4}{*}{APPNP} & Clean & 6.65\%           & 18.58\%          & 19.91\%          \\
                       & NIPA  & 9.90\%           & 21.15\%          & 20.14\%          \\
                       & AFGSM & 33.38\%          & 67.65\%          & \textbf{30.09\%} \\
                       & G-NIA & \textbf{99.85\%} & \textbf{99.14\%} & 28.91\% \\        
\bottomrule
\end{tabular}
\vspace{-5pt}
\end{table}

\begin{table}[]
\caption{Running time comparison on Single Target Attack on GCN on Reddit}
\label{tab:time}
\begin{tabular}{l|cccc}
\toprule
     & AFGSM & NIPA        & OPTI   & G-NIA   \\
\hline
Time (s) &  $2.72 \times 10^{-2}$   & $1.47\times 10^{-3}$  & 1.17 & $2.53\times 10^{-3}$ \\
\bottomrule
\end{tabular}
\vspace{-7pt}
\end{table}

\subsection{Efficiency Analysis}
The main reason that we require a generalizable model for the node injection attack is to avoid the repeating optimization for each attack and improve the efficiency of the attack.
To clarify the efficiency of our proposed method G-NIA, we report the running time of G-NIA, OPTI, and state-of-the-art baselines.
The other heuristic baselines are not included in the experiments, considering their poor attack performance.
The experiments are conducted on Single Target Attack on GCN on Reddit.
We profile the node injection attack on a server with GPU Tesla K80 and show the results in Table~\ref{tab:time}.
Note that running time refers to the average inference time of each node injection attack.

As shown in Figure~\ref{tab:time}, OPTI takes about 1.17 seconds to complete each attack, tens or even hundreds of times longer than other methods.
This is because that OPTI has to run a distinct optimization for each attack which is the reason why we need to further improve the attack efficiency.

Our proposed model G-NIA is significantly more efficient than OPTI.
For each attack, our model can generate the attributes and edges of the injected node in 0.00253 seconds, nearly 500 times faster than OPTI.
Moreover, G-NIA only takes 5 seconds to complete the node injection attacks on all test data, but OPTI needs 2333 seconds which is unbearable. 
In addition, G-NIA is 10 times faster than AFGSM. Because AFGSM has to loop many times to generate attributes, while G-NIA and NIPA can directly infer the results since they are both parametric models.
This gap will be even more significant as the number of attacks increases.
The results prove the efficiency of our G-NIA model when inferring.

\begin{figure}
\subfigure[Attributes visualization on Reddit]{
  \includegraphics[width=3.7cm]{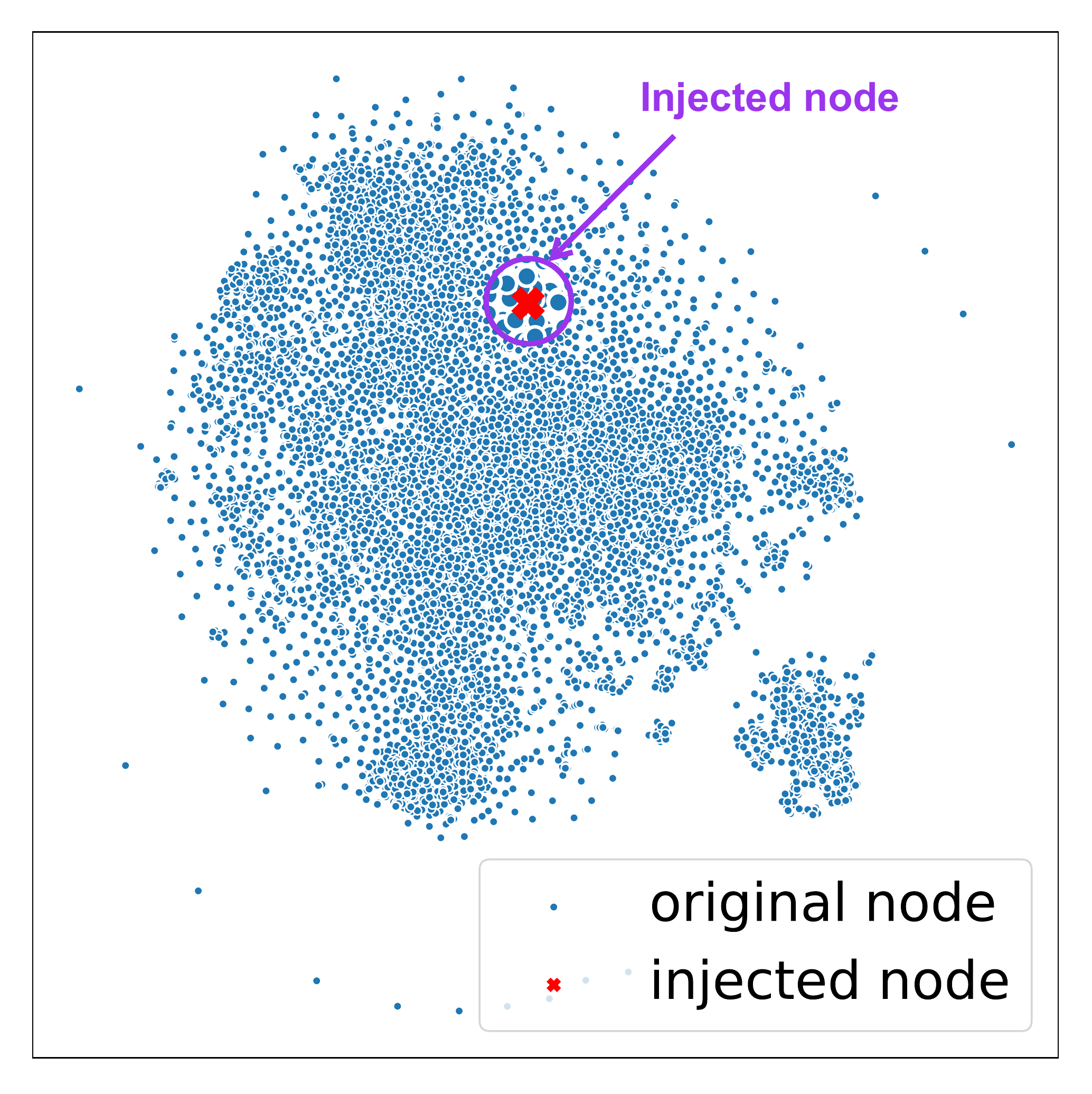}
 \includegraphics[width=3.7cm]{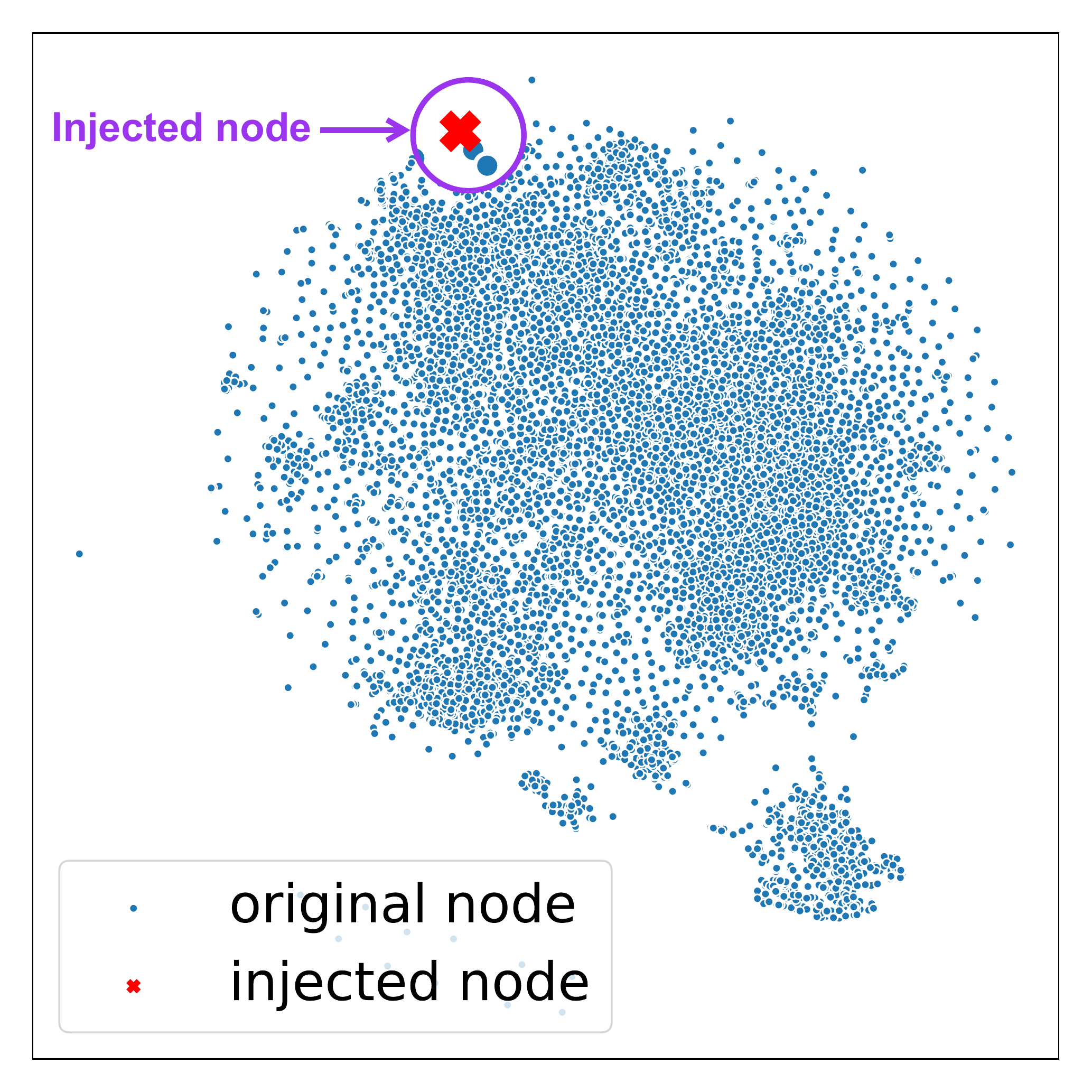}
\label{subfig:reddit}
}
\subfigure[Attributes visualization on ogbn-products]{
 \includegraphics[width=3.7cm]{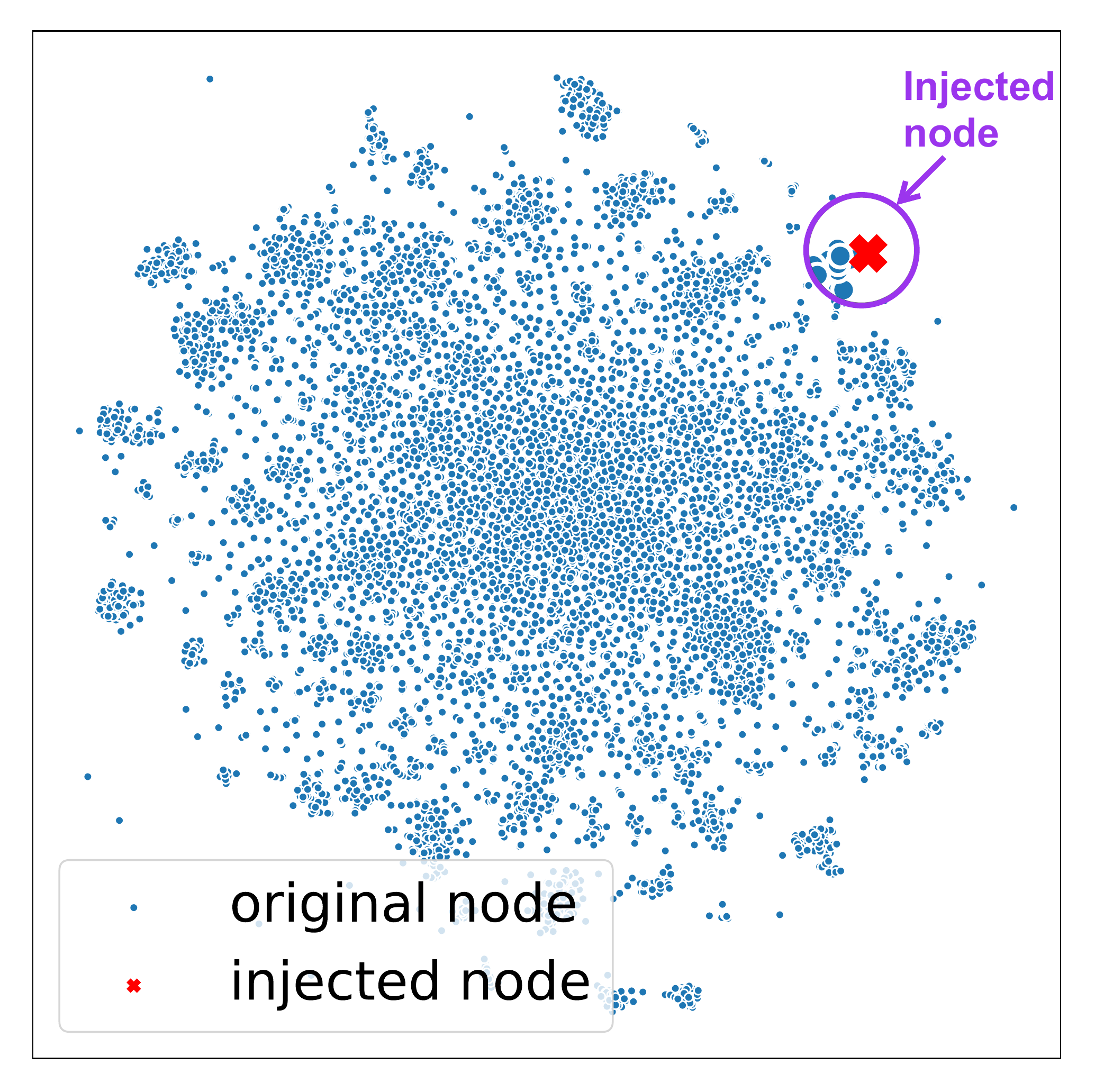}
 \includegraphics[width=3.7cm]{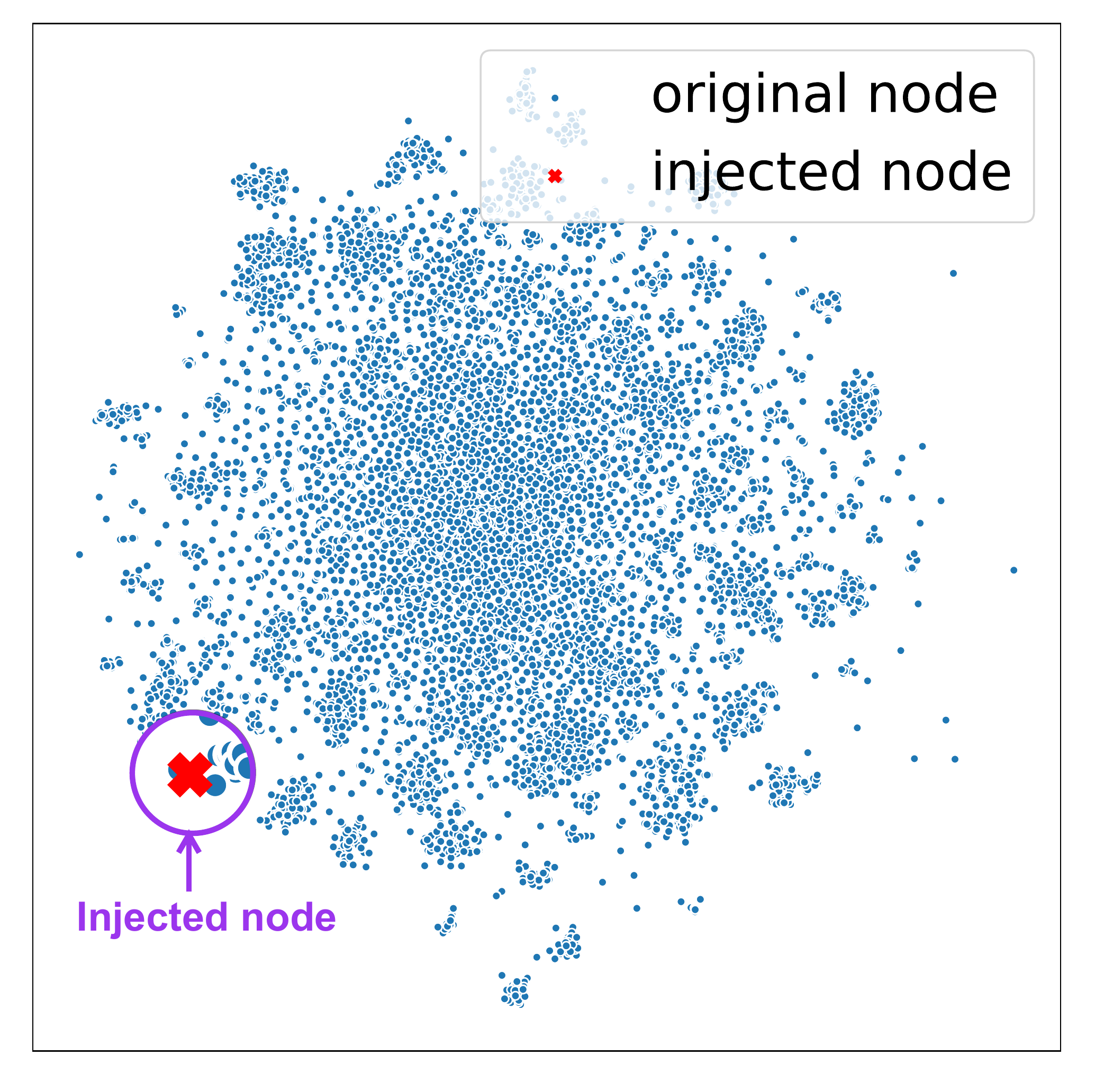}
 \label{subfig:ogb}
}
\caption{Case study on two continuous attributed graphs, Reddit and ogbn-products. The red cross indicates the injected node and the blue point refers to the original nodes.
} 
\label{fig:case_study}
\vspace{-8pt}
\end{figure}

\subsection{Case Study}

Our proposed G-NIA achieves great attack performances. 
Especially on two continuous graphs, the misclassification rate is close to 100\%.
We wonder whether it injects a weird node that is very different from the original ones.
Therefore, we visualize the attributes of the original nodes and the malicious node on Reddit and ogbn-products, to have an intuitive understanding of the malicious node injected by G-NIA.
We adopt the standard t-SNE technique~\cite{Maaten2008VisualizingDU} to visualize high-dimensional attributes.

The attribute visualizations on Reddit and ogbn-products are illustrated in Figure~\ref{fig:case_study}.
For each dataset, we show two attack cases on the Single Target Attack scenario.
Specifically, for each case, we randomly sample a target node and attack it via G-NIA.
In the four attack cases, the malicious node is mix up with the original nodes, further confirming that the single injected node is imperceptible and undetectable.
The results demonstrate that our attribute restriction and edge budget on the attack is helpful, and the injected node looks quite similar to existing nodes.

\subsection{Ablation Study}
We analyze the influence of each part of G-NIA through experiments and compare their attack performances when the corresponding part is removed.
Specifically, we remove attribute generation and edge generation respectively, and replace them with the Random strategy, namely G-NIA w/o Attr and G-NIA w/o Edge.
We also investigate the influence of jointly modeling by removing the attribute guidance for edges, i.e., G-NIA w/o Joint.  
That is, we replace  $\boldsymbol{r}_{inj}$ in the input of edge generation with an all-zero vector. 
Note that we take the experimental results of Single Target Attack on GCN on Reddit as an illustration.

\begin{figure}[t]
	\includegraphics[width=5.2cm]{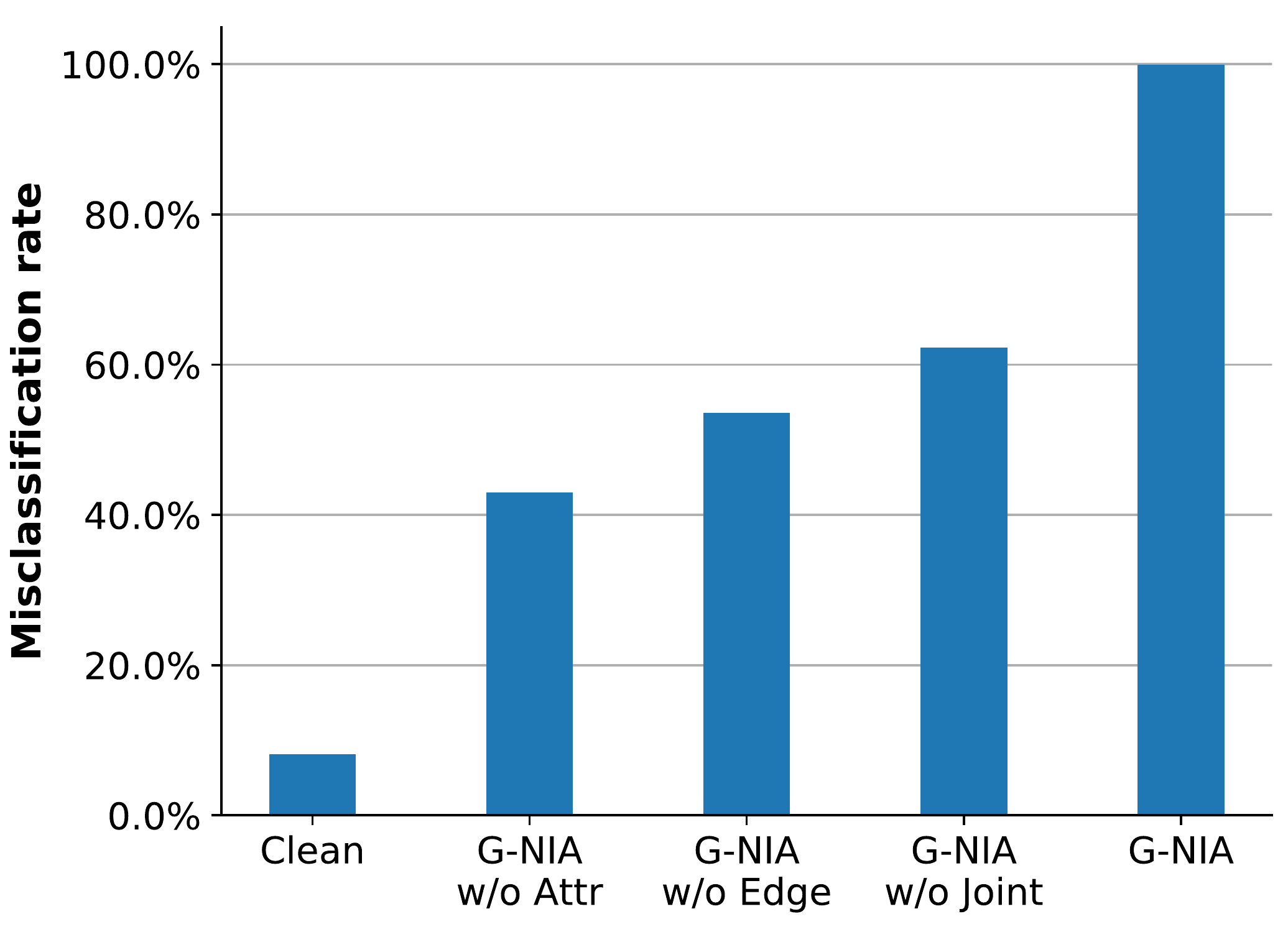}
	\caption{Ablation study. Misclassification rate on  clean graph and perturbed graphs attacked by G-NIA variants.}
	\label{fig:abl}
	\vspace{-8pt}
\end{figure}

As shown in Figure~\ref{fig:abl}, we find that the lack of any part damages the performance of G-NIA a lot. Among them, G-NIA w/o Attr performs worst, indicating that the most important part is attribute generation, which is consistent with our analysis in Section~\ref{sec:attr}.
G-NIA w/o Edge also loses half of the attack performance, implying the necessity of edge generation.
Besides, G-NIA w/o Joint only achieves a 62\% misclassification rate, which demonstrates the importance of jointly modeling. In other words, it makes sense to use malicious attributes to guide the generation of edges.
In a word, the experimental results prove that each part of G-NIA is necessary and helpful to improve the attack performance.

\section{Conclusions}
In this paper, we focus on an extremely limited scenario of single node injection evasion attack.
We first present an effective optimization-based approach (OPTI) to explore the upper bound of attack performance. 
However, the case-by-case optimization is not of practical significance, due to its computationally expensive cost.
To solve the dilemma, we propose a generalizable node injection attack model (G-NIA) to balance the attack performance and efficiency. 
G-NIA models the optimization process via a parametric model, to preserve the learned attack strategy and reuse it when inferring.
We conduct extensive attack experiments on three representative GNNs (GCN, GAT, APPNP) and three well-known datasets.
The experimental results reveal that G-NIA performs comparable with OPTI, but is much more efficient without re-optimization. 
With only injecting one node, G-NIA shows tremendous attack performances on Single Target Attack and Multiple Targets Attack tasks, outperforming state-of-the-art baselines. 
In addition, G-NIA can also carry out black-box attacks.

Although G-NIA is effective and efficient, there are still some remaining issues for future work. 
The first is whether the injected node is easy to be detected by some defense methods. If so, it is important to design and inject a more imperceptible malicious node.
The second is whether G-NIA is effective on the robust GNNs~\cite{Zgner2019CertifiableRA,jin2021robust,bojchevski_sparsesmoothing_2020} with adversarial training or robust training. We will explore these problems in the future.

\begin{acks}
This work is funded by the National Key R\&D Program of China (2020AAA0105200) and the National Natural Science Foundation of China under Grant Nos. 62102402, 91746301 and U1911401. Huawei Shen is also supported by Beijing Academy of Artificial Intelligence (BAAI) under the grant number BAAI2019QN0304 and K.C. Wong Education Foundation.
\end{acks}

\clearpage
\bibliographystyle{ACM-Reference-Format}
\bibliography{G-NIA}


\begin{thebibliography}{48}


\ifx \showCODEN    \undefined \def \showCODEN     #1{\unskip}     \fi
\ifx \showDOI      \undefined \def \showDOI       #1{#1}\fi
\ifx \showISBNx    \undefined \def \showISBNx     #1{\unskip}     \fi
\ifx \showISBNxiii \undefined \def \showISBNxiii  #1{\unskip}     \fi
\ifx \showISSN     \undefined \def \showISSN      #1{\unskip}     \fi
\ifx \showLCCN     \undefined \def \showLCCN      #1{\unskip}     \fi
\ifx \shownote     \undefined \def \shownote      #1{#1}          \fi
\ifx \showarticletitle \undefined \def \showarticletitle #1{#1}   \fi
\ifx \showURL      \undefined \def \showURL       {\relax}        \fi
\providecommand\bibfield[2]{#2}
\providecommand\bibinfo[2]{#2}
\providecommand\natexlab[1]{#1}
\providecommand\showeprint[2][]{arXiv:#2}

\bibitem[\protect\citeauthoryear{Akhtar and Mian}{Akhtar and Mian}{2018}]%
        {Akhtar2018ThreatOA}
\bibfield{author}{\bibinfo{person}{Naveed Akhtar} {and}
  \bibinfo{person}{Ajmal~S. Mian}.} \bibinfo{year}{2018}\natexlab{}.
\newblock \showarticletitle{Threat of Adversarial Attacks on Deep Learning in
  Computer Vision: A Survey}.
\newblock \bibinfo{journal}{\emph{IEEE Access}}  \bibinfo{volume}{6}
  (\bibinfo{year}{2018}), \bibinfo{pages}{14410--14430}.
\newblock


\bibitem[\protect\citeauthoryear{Baluja and Fischer}{Baluja and
  Fischer}{2018}]%
        {Baluja2018LearningTA}
\bibfield{author}{\bibinfo{person}{S. Baluja} {and} \bibinfo{person}{Ian~S.
  Fischer}.} \bibinfo{year}{2018}\natexlab{}.
\newblock \showarticletitle{Learning to Attack: Adversarial Transformation
  Networks}. In \bibinfo{booktitle}{\emph{Proceedings of the 30nd {AAAI}
  Conference on Artificial Intelligence}} \emph{(\bibinfo{series}{AAAI '18})}.
\newblock


\bibitem[\protect\citeauthoryear{Bojchevski and G{\"u}nnemann}{Bojchevski and
  G{\"u}nnemann}{2019}]%
        {Bojchevski2018AdversarialAO}
\bibfield{author}{\bibinfo{person}{Aleksandar Bojchevski} {and}
  \bibinfo{person}{Stephan G{\"u}nnemann}.} \bibinfo{year}{2019}\natexlab{}.
\newblock \showarticletitle{Adversarial Attacks on Node Embeddings via Graph
  Poisoning}. In \bibinfo{booktitle}{\emph{Proceedings of the 36th
  International Conference on Machine Learning}} \emph{(\bibinfo{series}{ICML
  '19})}. \bibinfo{pages}{695--704}.
\newblock


\bibitem[\protect\citeauthoryear{Bojchevski and G\"{u}nnemann}{Bojchevski and
  G\"{u}nnemann}{2019}]%
        {Bojchevski2019CertifiableRT}
\bibfield{author}{\bibinfo{person}{Aleksandar Bojchevski} {and}
  \bibinfo{person}{Stephan G\"{u}nnemann}.} \bibinfo{year}{2019}\natexlab{}.
\newblock \showarticletitle{Certifiable Robustness to Graph Perturbations}.
\newblock In \bibinfo{booktitle}{\emph{Advances in Neural Information
  Processing Systems 32}}. \bibinfo{pages}{8319--8330}.
\newblock


\bibitem[\protect\citeauthoryear{Bojchevski, Klicpera, and
  G{\"u}nnemann}{Bojchevski et~al\mbox{.}}{2020}]%
        {bojchevski_sparsesmoothing_2020}
\bibfield{author}{\bibinfo{person}{Aleksandar Bojchevski},
  \bibinfo{person}{Johannes Klicpera}, {and} \bibinfo{person}{Stephan
  G{\"u}nnemann}.} \bibinfo{year}{2020}\natexlab{}.
\newblock \showarticletitle{Efficient Robustness Certificates for Discrete
  Data: Sparsity-Aware Randomized Smoothing for Graphs, Images and More}. In
  \bibinfo{booktitle}{\emph{Proceedings of the 37th International Conference on
  Machine Learning}} \emph{(\bibinfo{series}{ICML '20})}.
  \bibinfo{pages}{11647--11657}.
\newblock


\bibitem[\protect\citeauthoryear{Cao, Shen, Gao, Wei, and Cheng}{Cao
  et~al\mbox{.}}{2020}]%
        {Cao2020PopularityPO}
\bibfield{author}{\bibinfo{person}{Qi Cao}, \bibinfo{person}{Huawei Shen},
  \bibinfo{person}{Jinhua Gao}, \bibinfo{person}{Bingzheng Wei}, {and}
  \bibinfo{person}{Xueqi Cheng}.} \bibinfo{year}{2020}\natexlab{}.
\newblock \showarticletitle{Popularity Prediction on Social Platforms with
  Coupled Graph Neural Networks}. In \bibinfo{booktitle}{\emph{Proceedings of
  the 13th International Conference on Web Search and Data Mining}}
  \emph{(\bibinfo{series}{WSDM '20})}. \bibinfo{pages}{70--78}.
\newblock


\bibitem[\protect\citeauthoryear{Chen, Li, Peng, Xie, Cao, Xu, He, and
  Zheng}{Chen et~al\mbox{.}}{2020}]%
        {Chen2020ASO}
\bibfield{author}{\bibinfo{person}{Liang Chen}, \bibinfo{person}{Jintang Li},
  \bibinfo{person}{Jiaying Peng}, \bibinfo{person}{Tao Xie},
  \bibinfo{person}{Zengxu Cao}, \bibinfo{person}{Kun Xu},
  \bibinfo{person}{Xiangnan He}, {and} \bibinfo{person}{Zibin Zheng}.}
  \bibinfo{year}{2020}\natexlab{}.
\newblock \showarticletitle{A Survey of Adversarial Learning on Graphs}.
\newblock \bibinfo{journal}{\emph{ArXiv}}  \bibinfo{volume}{abs/2003.05730}
  (\bibinfo{year}{2020}).
\newblock


\bibitem[\protect\citeauthoryear{Dai, Li, Tian, Huang, Wang, Zhu, and Song}{Dai
  et~al\mbox{.}}{2018}]%
        {Dai2018AdversarialAO}
\bibfield{author}{\bibinfo{person}{Hanjun Dai}, \bibinfo{person}{Hui Li},
  \bibinfo{person}{Tian Tian}, \bibinfo{person}{Xin Huang},
  \bibinfo{person}{Lin Wang}, \bibinfo{person}{Jun Zhu}, {and}
  \bibinfo{person}{Le Song}.} \bibinfo{year}{2018}\natexlab{}.
\newblock \showarticletitle{Adversarial Attack on Graph Structured Data}. In
  \bibinfo{booktitle}{\emph{Proceedings of the 35th International Conference on
  Machine Learning}} \emph{(\bibinfo{series}{ICML '18})}.
  \bibinfo{pages}{1123--1132}.
\newblock


\bibitem[\protect\citeauthoryear{Efraimidis and Spirakis}{Efraimidis and
  Spirakis}{2006}]%
        {Efraimidis2006WeightedRS}
\bibfield{author}{\bibinfo{person}{P. Efraimidis} {and} \bibinfo{person}{P.
  Spirakis}.} \bibinfo{year}{2006}\natexlab{}.
\newblock \showarticletitle{Weighted random sampling with a reservoir}.
\newblock \bibinfo{journal}{\emph{Inf. Process. Lett.}}  \bibinfo{volume}{97}
  (\bibinfo{year}{2006}), \bibinfo{pages}{181--185}.
\newblock


\bibitem[\protect\citeauthoryear{Fan, Ma, Li, He, Zhao, Tang, and Yin}{Fan
  et~al\mbox{.}}{2019}]%
        {fan2019graph}
\bibfield{author}{\bibinfo{person}{Wenqi Fan}, \bibinfo{person}{Yao Ma},
  \bibinfo{person}{Qing Li}, \bibinfo{person}{Yuan He}, \bibinfo{person}{Eric
  Zhao}, \bibinfo{person}{Jiliang Tang}, {and} \bibinfo{person}{Dawei Yin}.}
  \bibinfo{year}{2019}\natexlab{}.
\newblock \showarticletitle{Graph Neural Networks for Social Recommendation}.
  In \bibinfo{booktitle}{\emph{The World Wide Web Conference}}
  \emph{(\bibinfo{series}{WWW '19})}. \bibinfo{pages}{417--426}.
\newblock


\bibitem[\protect\citeauthoryear{Finkelshtein, Baskin, Zheltonozhskii, and
  Alon}{Finkelshtein et~al\mbox{.}}{2020}]%
        {Finkelshtein2020SingleNodeAF}
\bibfield{author}{\bibinfo{person}{Ben Finkelshtein}, \bibinfo{person}{Chaim
  Baskin}, \bibinfo{person}{Evgenii Zheltonozhskii}, {and} \bibinfo{person}{Uri
  Alon}.} \bibinfo{year}{2020}\natexlab{}.
\newblock \showarticletitle{Single-Node Attack for Fooling Graph Neural
  Networks}.
\newblock \bibinfo{journal}{\emph{ArXiv}}  \bibinfo{volume}{abs/2011.03574}
  (\bibinfo{year}{2020}).
\newblock


\bibitem[\protect\citeauthoryear{Graves}{Graves}{2013}]%
        {Graves2013GeneratingSW}
\bibfield{author}{\bibinfo{person}{A. Graves}.}
  \bibinfo{year}{2013}\natexlab{}.
\newblock \showarticletitle{Generating Sequences With Recurrent Neural
  Networks}.
\newblock \bibinfo{journal}{\emph{ArXiv}}  \bibinfo{volume}{abs/1308.0850}
  (\bibinfo{year}{2013}).
\newblock


\bibitem[\protect\citeauthoryear{Gumbel}{Gumbel}{1954}]%
        {gumbel1954statistical}
\bibfield{author}{\bibinfo{person}{Emil~Julius Gumbel}.}
  \bibinfo{year}{1954}\natexlab{}.
\newblock \bibinfo{booktitle}{\emph{Statistical theory of extreme values and
  some practical applications: a series of lectures}}.
  Vol.~\bibinfo{volume}{33}.
\newblock \bibinfo{publisher}{US Government Printing Office}.
\newblock


\bibitem[\protect\citeauthoryear{Hamilton, Ying, and Leskovec}{Hamilton
  et~al\mbox{.}}{2017}]%
        {hamilton2017inductive}
\bibfield{author}{\bibinfo{person}{William Hamilton}, \bibinfo{person}{Zhitao
  Ying}, {and} \bibinfo{person}{Jure Leskovec}.}
  \bibinfo{year}{2017}\natexlab{}.
\newblock \showarticletitle{Inductive Representation Learning on Large Graphs}.
\newblock In \bibinfo{booktitle}{\emph{Advances in Neural Information
  Processing Systems 30}}. \bibinfo{pages}{1024--1034}.
\newblock


\bibitem[\protect\citeauthoryear{Hu, Fey, Zitnik, Dong, Ren, Liu, Catasta, and
  Leskovec}{Hu et~al\mbox{.}}{2020}]%
        {Hu2020OGB}
\bibfield{author}{\bibinfo{person}{Weihua Hu}, \bibinfo{person}{Matthias Fey},
  \bibinfo{person}{Marinka Zitnik}, \bibinfo{person}{Yuxiao Dong},
  \bibinfo{person}{Hongyu Ren}, \bibinfo{person}{Bowen Liu},
  \bibinfo{person}{Michele Catasta}, {and} \bibinfo{person}{Jure Leskovec}.}
  \bibinfo{year}{2020}\natexlab{}.
\newblock \showarticletitle{Open Graph Benchmark: Datasets for Machine Learning
  on Graphs}. In \bibinfo{booktitle}{\emph{Advances in Neural Information
  Processing Systems 33}}.
\newblock


\bibitem[\protect\citeauthoryear{Jia, Wang, Cao, and Gong}{Jia
  et~al\mbox{.}}{2020}]%
        {Jia2020CertifiedRO}
\bibfield{author}{\bibinfo{person}{Jinyuan Jia}, \bibinfo{person}{Binghui
  Wang}, \bibinfo{person}{Xiaoyu Cao}, {and} \bibinfo{person}{Neil~Zhenqiang
  Gong}.} \bibinfo{year}{2020}\natexlab{}.
\newblock \showarticletitle{Certified Robustness of Community Detection against
  Adversarial Structural Perturbation via Randomized Smoothing}. In
  \bibinfo{booktitle}{\emph{Proceedings of The Web Conference 2020}}
  \emph{(\bibinfo{series}{WWW '20})}. \bibinfo{pages}{2718--2724}.
\newblock


\bibitem[\protect\citeauthoryear{Jin and Zhang}{Jin and Zhang}{2021}]%
        {jin2021robust}
\bibfield{author}{\bibinfo{person}{Hongwei Jin} {and} \bibinfo{person}{Xinhua
  Zhang}.} \bibinfo{year}{2021}\natexlab{}.
\newblock \showarticletitle{Robust Training of Graph Convolutional Networks via
  Latent Perturbation}. In \bibinfo{booktitle}{\emph{Machine Learning and
  Knowledge Discovery in Databases: European Conference}}
  \emph{(\bibinfo{series}{ECML/PKDD})}. \bibinfo{pages}{394--411}.
\newblock


\bibitem[\protect\citeauthoryear{Jin, Li, Xu, Wang, and Tang}{Jin
  et~al\mbox{.}}{2020}]%
        {Jin2020AdversarialAA}
\bibfield{author}{\bibinfo{person}{Wei Jin}, \bibinfo{person}{Yaxin Li},
  \bibinfo{person}{Han Xu}, \bibinfo{person}{Yiqi Wang}, {and}
  \bibinfo{person}{Jiliang Tang}.} \bibinfo{year}{2020}\natexlab{}.
\newblock \showarticletitle{Adversarial Attacks and Defenses on Graphs: A
  Review and Empirical Study}.
\newblock \bibinfo{journal}{\emph{ArXiv}}  \bibinfo{volume}{abs/2003.00653}
  (\bibinfo{year}{2020}).
\newblock


\bibitem[\protect\citeauthoryear{Kipf and Welling}{Kipf and Welling}{2017}]%
        {kipf2017semi}
\bibfield{author}{\bibinfo{person}{Thomas~N. Kipf} {and} \bibinfo{person}{Max
  Welling}.} \bibinfo{year}{2017}\natexlab{}.
\newblock \showarticletitle{Semi-Supervised Classification with Graph
  Convolutional Networks}. In \bibinfo{booktitle}{\emph{International
  Conference on Learning Representations}} \emph{(\bibinfo{series}{ICLR '17})}.
\newblock


\bibitem[\protect\citeauthoryear{Klicpera, Bojchevski, and
  G{\"u}nnemann}{Klicpera et~al\mbox{.}}{2019}]%
        {Klicpera2018PredictTP}
\bibfield{author}{\bibinfo{person}{Johannes Klicpera},
  \bibinfo{person}{Aleksandar Bojchevski}, {and} \bibinfo{person}{Stephan
  G{\"u}nnemann}.} \bibinfo{year}{2019}\natexlab{}.
\newblock \showarticletitle{Predict then Propagate: Graph Neural Networks meet
  Personalized PageRank}. In \bibinfo{booktitle}{\emph{International Conference
  on Learning Representations}} \emph{(\bibinfo{series}{ICLR '19})}.
\newblock


\bibitem[\protect\citeauthoryear{K{\"u}gler, Distergoft, Kuijper, and
  Mukhopadhyay}{K{\"u}gler et~al\mbox{.}}{2018}]%
        {Kgler2018ExploringAE}
\bibfield{author}{\bibinfo{person}{David K{\"u}gler},
  \bibinfo{person}{Alexander Distergoft}, \bibinfo{person}{Arjan Kuijper},
  {and} \bibinfo{person}{A. Mukhopadhyay}.} \bibinfo{year}{2018}\natexlab{}.
\newblock \showarticletitle{Exploring Adversarial Examples: Patterns of
  One-Pixel Attacks}. In \bibinfo{booktitle}{\emph{MLCN/DLF/iMIMIC@MICCAI}}.
\newblock


\bibitem[\protect\citeauthoryear{Lin, Zhou, Yang, Wu, Wang, Cao, and Wang}{Lin
  et~al\mbox{.}}{2020}]%
        {Lin2020ExploratoryAA}
\bibfield{author}{\bibinfo{person}{Xixun Lin}, \bibinfo{person}{C. Zhou},
  \bibinfo{person}{H. Yang}, \bibinfo{person}{Jia Wu}, \bibinfo{person}{Haibo
  Wang}, \bibinfo{person}{Yanan Cao}, {and} \bibinfo{person}{Bin Wang}.}
  \bibinfo{year}{2020}\natexlab{}.
\newblock \showarticletitle{Exploratory Adversarial Attacks on Graph Neural
  Networks}. In \bibinfo{booktitle}{\emph{20th {IEEE} International Conference
  on Data Mining}} \emph{(\bibinfo{series}{ICDM '20})}.
  \bibinfo{pages}{791--800}.
\newblock


\bibitem[\protect\citeauthoryear{Ma, Liu, Zhao, Liu, Tang, and Shah}{Ma
  et~al\mbox{.}}{2020}]%
        {Ma2020AUV}
\bibfield{author}{\bibinfo{person}{Yao Ma}, \bibinfo{person}{Xiaorui Liu},
  \bibinfo{person}{Tong Zhao}, \bibinfo{person}{Yozen Liu},
  \bibinfo{person}{Jiliang Tang}, {and} \bibinfo{person}{Neil Shah}.}
  \bibinfo{year}{2020}\natexlab{}.
\newblock \showarticletitle{A Unified View on Graph Neural Networks as Graph
  Signal Denoising}.
\newblock \bibinfo{journal}{\emph{ArXiv}}  \bibinfo{volume}{abs/2010.01777}
  (\bibinfo{year}{2020}).
\newblock


\bibitem[\protect\citeauthoryear{Ma, Wang, Derr, Wu, and Tang}{Ma
  et~al\mbox{.}}{2021}]%
        {Ma2021GraphRewir}
\bibfield{author}{\bibinfo{person}{Yao Ma}, \bibinfo{person}{Suhang Wang},
  \bibinfo{person}{Tyler Derr}, \bibinfo{person}{Lingfei Wu}, {and}
  \bibinfo{person}{Jiliang Tang}.} \bibinfo{year}{2021}\natexlab{}.
\newblock \showarticletitle{Graph Adversarial Attack via Rewiring}. In
  \bibinfo{booktitle}{\emph{Proceedings of the 27th ACM SIGKDD International
  Conference on Knowledge Discovery \& Data Mining}}
  \emph{(\bibinfo{series}{KDD '21})}. \bibinfo{pages}{1161--1169}.
\newblock


\bibitem[\protect\citeauthoryear{Maaten and Hinton}{Maaten and Hinton}{2008}]%
        {Maaten2008VisualizingDU}
\bibfield{author}{\bibinfo{person}{L.~V.~D. Maaten} {and}
  \bibinfo{person}{Geoffrey~E. Hinton}.} \bibinfo{year}{2008}\natexlab{}.
\newblock \showarticletitle{Visualizing Data using t-SNE}.
\newblock \bibinfo{journal}{\emph{Journal of Machine Learning Research}}
  \bibinfo{volume}{9} (\bibinfo{year}{2008}), \bibinfo{pages}{2579--2605}.
\newblock


\bibitem[\protect\citeauthoryear{Qiu, Tang, Ma, Dong, Wang, and Tang}{Qiu
  et~al\mbox{.}}{2018}]%
        {Qiu2018DeepInfSI}
\bibfield{author}{\bibinfo{person}{Jiezhong Qiu}, \bibinfo{person}{Jian Tang},
  \bibinfo{person}{Hao Ma}, \bibinfo{person}{Yuxiao Dong},
  \bibinfo{person}{Kuansan Wang}, {and} \bibinfo{person}{Jie Tang}.}
  \bibinfo{year}{2018}\natexlab{}.
\newblock \showarticletitle{DeepInf: Social Influence Prediction with Deep
  Learning}. In \bibinfo{booktitle}{\emph{Proceedings of the 24th ACM SIGKDD
  International Conference on Knowledge Discovery \& Data Mining}}
  \emph{(\bibinfo{series}{KDD '18})}. \bibinfo{pages}{2110--2119}.
\newblock


\bibitem[\protect\citeauthoryear{Ribeiro, Saverese, and Figueiredo}{Ribeiro
  et~al\mbox{.}}{2017}]%
        {10.1145/3097983.3098061}
\bibfield{author}{\bibinfo{person}{Leonardo~F.R. Ribeiro},
  \bibinfo{person}{Pedro~H.P. Saverese}, {and} \bibinfo{person}{Daniel~R.
  Figueiredo}.} \bibinfo{year}{2017}\natexlab{}.
\newblock \showarticletitle{Struc2vec: Learning Node Representations from
  Structural Identity}. In \bibinfo{booktitle}{\emph{Proceedings of the 23rd
  ACM SIGKDD International Conference on Knowledge Discovery \& Data Mining}}
  \emph{(\bibinfo{series}{KDD '17})}. \bibinfo{pages}{385--394}.
\newblock


\bibitem[\protect\citeauthoryear{Rieck, Bock, and Borgwardt}{Rieck
  et~al\mbox{.}}{2019}]%
        {rieck2019persistent}
\bibfield{author}{\bibinfo{person}{Bastian Rieck}, \bibinfo{person}{Christian
  Bock}, {and} \bibinfo{person}{Karsten Borgwardt}.}
  \bibinfo{year}{2019}\natexlab{}.
\newblock \showarticletitle{A persistent weisfeiler-lehman procedure for graph
  classification}. In \bibinfo{booktitle}{\emph{Proceedings of the 36th
  International Conference on Machine Learning}} \emph{(\bibinfo{series}{ICML
  '19})}. \bibinfo{pages}{5448--5458}.
\newblock


\bibitem[\protect\citeauthoryear{Su, Vargas, and Sakurai}{Su
  et~al\mbox{.}}{2019}]%
        {Su2019OnePA}
\bibfield{author}{\bibinfo{person}{Jiawei Su},
  \bibinfo{person}{Danilo~Vasconcellos Vargas}, {and} \bibinfo{person}{K.
  Sakurai}.} \bibinfo{year}{2019}\natexlab{}.
\newblock \showarticletitle{One Pixel Attack for Fooling Deep Neural Networks}.
\newblock \bibinfo{journal}{\emph{IEEE Transactions on Evolutionary
  Computation}}  \bibinfo{volume}{23} (\bibinfo{year}{2019}),
  \bibinfo{pages}{828--841}.
\newblock


\bibitem[\protect\citeauthoryear{Sun, Wang, Yu, and Li}{Sun
  et~al\mbox{.}}{2018}]%
        {Sun2018AdversarialAA}
\bibfield{author}{\bibinfo{person}{Lichao Sun}, \bibinfo{person}{Ji Wang},
  \bibinfo{person}{Philip~S. Yu}, {and} \bibinfo{person}{Bo Li}.}
  \bibinfo{year}{2018}\natexlab{}.
\newblock \showarticletitle{Adversarial Attack and Defense on Graph Data: A
  Survey}.
\newblock \bibinfo{journal}{\emph{ArXiv}}  \bibinfo{volume}{abs/1812.10528}
  (\bibinfo{year}{2018}).
\newblock


\bibitem[\protect\citeauthoryear{Sun, Wang, Tang, Hsieh, and Honavar}{Sun
  et~al\mbox{.}}{2020}]%
        {Sun2020AdversarialAO}
\bibfield{author}{\bibinfo{person}{Yiwei Sun}, \bibinfo{person}{Suhang Wang},
  \bibinfo{person}{Xian-Feng Tang}, \bibinfo{person}{Tsung-Yu Hsieh}, {and}
  \bibinfo{person}{Vasant~G Honavar}.} \bibinfo{year}{2020}\natexlab{}.
\newblock \showarticletitle{Adversarial Attacks on Graph Neural Networks via
  Node Injections: A Hierarchical Reinforcement Learning Approach}. In
  \bibinfo{booktitle}{\emph{Proceedings of The Web Conference 2020}}
  \emph{(\bibinfo{series}{WWW '20})}. \bibinfo{pages}{673--683}.
\newblock


\bibitem[\protect\citeauthoryear{Tao, Shen, Cao, Hou, and Cheng.}{Tao
  et~al\mbox{.}}{2021}]%
        {tao2021advimmune}
\bibfield{author}{\bibinfo{person}{Shuchang Tao}, \bibinfo{person}{Huawei
  Shen}, \bibinfo{person}{Qi Cao}, \bibinfo{person}{Liang Hou}, {and}
  \bibinfo{person}{Xueqi Cheng.}} \bibinfo{year}{2021}\natexlab{}.
\newblock \showarticletitle{Adversarial Immunization for Certifiable Robustness
  on Graphs}. In \bibinfo{booktitle}{\emph{Proceedings of the 14th ACM
  International Conference on Web Search and Data Mining}}
  \emph{(\bibinfo{series}{WSDM'21})}. \bibinfo{pages}{698--706}.
\newblock


\bibitem[\protect\citeauthoryear{Vargas and Su}{Vargas and Su}{2020}]%
        {Vargas2020UnderstandingTO}
\bibfield{author}{\bibinfo{person}{Danilo~Vasconcellos Vargas} {and}
  \bibinfo{person}{Jiawei Su}.} \bibinfo{year}{2020}\natexlab{}.
\newblock \showarticletitle{Understanding the One Pixel Attack: Propagation
  Maps and Locality Analysis}.
\newblock \bibinfo{journal}{\emph{ArXiv}}  \bibinfo{volume}{abs/1902.02947}
  (\bibinfo{year}{2020}).
\newblock


\bibitem[\protect\citeauthoryear{Veli{\v{c}}kovi{\'{c}}, Cucurull, Casanova,
  Romero, Li{\`{o}}, and Bengio}{Veli{\v{c}}kovi{\'{c}} et~al\mbox{.}}{2018}]%
        {velickovic2018graph}
\bibfield{author}{\bibinfo{person}{Petar Veli{\v{c}}kovi{\'{c}}},
  \bibinfo{person}{Guillem Cucurull}, \bibinfo{person}{Arantxa Casanova},
  \bibinfo{person}{Adriana Romero}, \bibinfo{person}{Pietro Li{\`{o}}}, {and}
  \bibinfo{person}{Yoshua Bengio}.} \bibinfo{year}{2018}\natexlab{}.
\newblock \showarticletitle{{Graph Attention Networks}}. In
  \bibinfo{booktitle}{\emph{International Conference on Learning
  Representations}} \emph{(\bibinfo{series}{ICLR '18})}.
\newblock


\bibitem[\protect\citeauthoryear{Wang and Gong}{Wang and Gong}{2019}]%
        {Wang2019AttackingGC}
\bibfield{author}{\bibinfo{person}{Binghui Wang} {and}
  \bibinfo{person}{Neil~Zhenqiang Gong}.} \bibinfo{year}{2019}\natexlab{}.
\newblock \showarticletitle{Attacking Graph-Based Classification via
  Manipulating the Graph Structure}. In \bibinfo{booktitle}{\emph{Proceedings
  of the 2019 ACM SIGSAC Conference on Computer and Communications Security}}
  \emph{(\bibinfo{series}{CCS '19})}. \bibinfo{pages}{2023--2040}.
\newblock


\bibitem[\protect\citeauthoryear{Wang, Luo, Suya, Li, Yang, and Zheng}{Wang
  et~al\mbox{.}}{2020a}]%
        {Wang2020ScalableAO}
\bibfield{author}{\bibinfo{person}{Jihong Wang}, \bibinfo{person}{Minnan Luo},
  \bibinfo{person}{Fnu Suya}, \bibinfo{person}{Jundong Li},
  \bibinfo{person}{Zijiang Yang}, {and} \bibinfo{person}{Qinghua Zheng}.}
  \bibinfo{year}{2020}\natexlab{a}.
\newblock \showarticletitle{Scalable Attack on Graph Data by Injecting Vicious
  Nodes}.
\newblock \bibinfo{journal}{\emph{ArXiv}}  \bibinfo{volume}{abs/2004.13825}
  (\bibinfo{year}{2020}).
\newblock


\bibitem[\protect\citeauthoryear{Wang, Zhu, Bo, Cui, Shi, and Pei}{Wang
  et~al\mbox{.}}{2020b}]%
        {wang2020amgcn}
\bibfield{author}{\bibinfo{person}{Xiao Wang}, \bibinfo{person}{Meiqi Zhu},
  \bibinfo{person}{Deyu Bo}, \bibinfo{person}{Peng Cui}, \bibinfo{person}{Chuan
  Shi}, {and} \bibinfo{person}{Jian Pei}.} \bibinfo{year}{2020}\natexlab{b}.
\newblock \showarticletitle{AM-GCN: Adaptive Multi-Channel Graph Convolutional
  Networks}. In \bibinfo{booktitle}{\emph{Proceedings of the 26th ACM SIGKDD
  International Conference on Knowledge Discovery \& Data Mining}}
  \emph{(\bibinfo{series}{KDD '20})}. \bibinfo{pages}{1243–1253}.
\newblock


\bibitem[\protect\citeauthoryear{Xu, Shen, Cao, Cen, and Cheng}{Xu
  et~al\mbox{.}}{2019b}]%
        {xu2019graphheat}
\bibfield{author}{\bibinfo{person}{Bingbing Xu}, \bibinfo{person}{Huawei Shen},
  \bibinfo{person}{Qi Cao}, \bibinfo{person}{Keting Cen}, {and}
  \bibinfo{person}{Xueqi Cheng}.} \bibinfo{year}{2019}\natexlab{b}.
\newblock \showarticletitle{Graph convolutional networks using heat kernel for
  semi-supervised learning}. In \bibinfo{booktitle}{\emph{Proceedings of the
  28th International Joint Conference on Artificial Intelligence}}
  \emph{(\bibinfo{series}{IJCAI '19})}. \bibinfo{pages}{1928--1934}.
\newblock


\bibitem[\protect\citeauthoryear{Xu, Shen, Cao, Qiu, and Cheng}{Xu
  et~al\mbox{.}}{2019c}]%
        {xu2018gwnn}
\bibfield{author}{\bibinfo{person}{Bingbing Xu}, \bibinfo{person}{Huawei Shen},
  \bibinfo{person}{Qi Cao}, \bibinfo{person}{Yunqi Qiu}, {and}
  \bibinfo{person}{Xueqi Cheng}.} \bibinfo{year}{2019}\natexlab{c}.
\newblock \showarticletitle{Graph Wavelet Neural Network}. In
  \bibinfo{booktitle}{\emph{International Conference on Learning
  Representations}} \emph{(\bibinfo{series}{ICLR '19})}.
\newblock


\bibitem[\protect\citeauthoryear{Xu, Ma, Liu, Deb, Liu, Tang, and Jain}{Xu
  et~al\mbox{.}}{2020}]%
        {XuAADIG}
\bibfield{author}{\bibinfo{person}{Han Xu}, \bibinfo{person}{Yao Ma},
  \bibinfo{person}{Haochen Liu}, \bibinfo{person}{Debayan Deb},
  \bibinfo{person}{Hui Liu}, \bibinfo{person}{Jiliang Tang}, {and}
  \bibinfo{person}{Anil~K. Jain}.} \bibinfo{year}{2020}\natexlab{}.
\newblock \showarticletitle{Adversarial Attacks and Defenses in Images, Graphs
  and Text: {A} Review}.
\newblock \bibinfo{journal}{\emph{Int. J. Autom. Comput.}}
  \bibinfo{volume}{17}, \bibinfo{number}{2} (\bibinfo{year}{2020}),
  \bibinfo{pages}{151--178}.
\newblock


\bibitem[\protect\citeauthoryear{Xu, Hu, Leskovec, and Jegelka}{Xu
  et~al\mbox{.}}{2019a}]%
        {Xu2019HowPA}
\bibfield{author}{\bibinfo{person}{Keyulu Xu}, \bibinfo{person}{Weihua Hu},
  \bibinfo{person}{J. Leskovec}, {and} \bibinfo{person}{S. Jegelka}.}
  \bibinfo{year}{2019}\natexlab{a}.
\newblock \showarticletitle{How Powerful are Graph Neural Networks?}. In
  \bibinfo{booktitle}{\emph{International Conference on Learning
  Representations}} \emph{(\bibinfo{series}{ICLR '19})}.
\newblock


\bibitem[\protect\citeauthoryear{You, Gomes{-}Selman, Ying, and Leskovec}{You
  et~al\mbox{.}}{2021}]%
        {You2021IDGNN}
\bibfield{author}{\bibinfo{person}{Jiaxuan You}, \bibinfo{person}{Jonathan
  Gomes{-}Selman}, \bibinfo{person}{Rex Ying}, {and} \bibinfo{person}{Jure
  Leskovec}.} \bibinfo{year}{2021}\natexlab{}.
\newblock \showarticletitle{Identity-aware Graph Neural Networks}. In
  \bibinfo{booktitle}{\emph{Proceedings of the 33nd {AAAI} Conference on
  Artificial Intelligence}} \emph{(\bibinfo{series}{AAAI '21})}.
\newblock


\bibitem[\protect\citeauthoryear{Zeng, Zhou, Srivastava, Kannan, and
  Prasanna}{Zeng et~al\mbox{.}}{2020}]%
        {Zeng2019GraphSAINTGS}
\bibfield{author}{\bibinfo{person}{Hanqing Zeng}, \bibinfo{person}{Hongkuan
  Zhou}, \bibinfo{person}{Ajitesh Srivastava}, \bibinfo{person}{Rajgopal
  Kannan}, {and} \bibinfo{person}{Viktor Prasanna}.}
  \bibinfo{year}{2020}\natexlab{}.
\newblock \showarticletitle{{GraphSAINT}: Graph Sampling Based Inductive
  Learning Method}. In \bibinfo{booktitle}{\emph{International Conference on
  Learning Representations}} \emph{(\bibinfo{series}{ICLR '20})}.
\newblock


\bibitem[\protect\citeauthoryear{Zhang, Hu, Shi, and Wang}{Zhang
  et~al\mbox{.}}{2020}]%
        {ZhangHSW20}
\bibfield{author}{\bibinfo{person}{Mengmei Zhang}, \bibinfo{person}{Linmei Hu},
  \bibinfo{person}{Chuan Shi}, {and} \bibinfo{person}{Xiao Wang}.}
  \bibinfo{year}{2020}\natexlab{}.
\newblock \showarticletitle{Adversarial Label-Flipping Attack and Defense for
  Graph Neural Networks}. In \bibinfo{booktitle}{\emph{20th {IEEE}
  International Conference on Data Mining}} \emph{(\bibinfo{series}{ICDM
  '20})}. \bibinfo{pages}{791--800}.
\newblock


\bibitem[\protect\citeauthoryear{Zhang, Khan, and Coates}{Zhang
  et~al\mbox{.}}{[n.d.]}]%
        {zhang2019comparing}
\bibfield{author}{\bibinfo{person}{Yingxue Zhang}, \bibinfo{person}{S Khan},
  {and} \bibinfo{person}{Mark Coates}.} \bibinfo{year}{[n.d.]}\natexlab{}.
\newblock \showarticletitle{Comparing and detecting adversarial attacks for
  graph deep learning}. In \bibinfo{booktitle}{\emph{Proc. Representation
  Learning on Graphs and Manifolds Workshop, Int. Conf. Learning
  Representations, New Orleans, LA, USA}} \emph{(\bibinfo{series}{RLGM @ ICLR
  '19})}.
\newblock


\bibitem[\protect\citeauthoryear{Z\"{u}gner, Akbarnejad, and
  G\"{u}nnemann}{Z\"{u}gner et~al\mbox{.}}{2018}]%
        {zugner2018adversarial}
\bibfield{author}{\bibinfo{person}{Daniel Z\"{u}gner}, \bibinfo{person}{Amir
  Akbarnejad}, {and} \bibinfo{person}{Stephan G\"{u}nnemann}.}
  \bibinfo{year}{2018}\natexlab{}.
\newblock \showarticletitle{Adversarial attacks on neural networks for graph
  data}. In \bibinfo{booktitle}{\emph{Proceedings of the 24th ACM SIGKDD
  International Conference on Knowledge Discovery \& Data Mining}}
  \emph{(\bibinfo{series}{KDD '18})}. \bibinfo{pages}{2847--2856}.
\newblock


\bibitem[\protect\citeauthoryear{Z{\"u}gner and G{\"u}nnemann}{Z{\"u}gner and
  G{\"u}nnemann}{2019a}]%
        {zugner_adversarial_2019}
\bibfield{author}{\bibinfo{person}{Daniel Z{\"u}gner} {and}
  \bibinfo{person}{Stephan G{\"u}nnemann}.} \bibinfo{year}{2019}\natexlab{a}.
\newblock \showarticletitle{Adversarial Attacks on Graph Neural Networks via
  Meta Learning}. In \bibinfo{booktitle}{\emph{International Conference on
  Learning Representations}} \emph{(\bibinfo{series}{ICLR '19})}.
\newblock


\bibitem[\protect\citeauthoryear{Z{\"u}gner and G{\"u}nnemann}{Z{\"u}gner and
  G{\"u}nnemann}{2019b}]%
        {Zgner2019CertifiableRA}
\bibfield{author}{\bibinfo{person}{Daniel Z{\"u}gner} {and}
  \bibinfo{person}{Stephan G{\"u}nnemann}.} \bibinfo{year}{2019}\natexlab{b}.
\newblock \showarticletitle{Certifiable Robustness and Robust Training for
  Graph Convolutional Networks}. In \bibinfo{booktitle}{\emph{Proceedings of
  the 25th ACM SIGKDD International Conference on Knowledge Discovery \& Data
  Mining}} \emph{(\bibinfo{series}{KDD '19})}. \bibinfo{pages}{246--256}.
\newblock


\end{thebibliography}

\end{document}